\documentclass[lettersize,journal]{IEEEtran}
\usepackage{amsmath,amsfonts}
\usepackage{algorithmic}
\usepackage{algorithm}
\usepackage{array}
\usepackage[caption=false,font=footnotesize,labelfont=rm,textfont=rm]{subfig}
\usepackage{textcomp}
\usepackage{stfloats}
\usepackage{picture}
\usepackage{url}
\usepackage{verbatim}
\usepackage{graphicx}
\usepackage{cite}
\usepackage{graphicx}
\usepackage{pgfplots}
\usepackage{booktabs}
\usepackage{multicol}
\usepackage{multirow}
\usepackage{xcolor}
\usepackage{color} 
\usepackage{soul}
\soulregister{\cite}7 
\soulregister{\citep}7 
\soulregister{\citet}7 
\soulregister{\ref}7 
\soulregister{\pageref}7 

\usepackage{tikz}
\usepackage[implicit=false]{hyperref}
\hypersetup{hidelinks,
	colorlinks=true,
	allcolors=black,
	pdfstartview=Fit,
	breaklinks=true}
\definecolor{lime}{HTML}{A6CE39}
\DeclareRobustCommand{\orcidicon}{
\hspace{-3mm}
\begin{tikzpicture}
\draw[lime, fill=lime] (0,0)
circle[radius=0.12]
node[white]{{\fontfamily{qag}\selectfont \tiny \.{I}D}};
\end{tikzpicture}
\hspace{-3mm}
}
\foreach \x in {GT, WYB, ZKH, CP, CT}{
\expandafter\xdef\csname orcid\x\endcsname{\noexpand\href{https://orcid.org/\csname orcidauthor\x\endcsname}{\noexpand\orcidicon}}
}

\newcommand\etc{\textit{etc}{\@}}
\newcommand\ie{\textit{e.g.\ }{}{\@}}

\begin{document}
 
\title{Towards an Effective and Efficient Transformer for Rain-by-snow Weather Removal

\thanks{The research is partially supported by the National Key R \& D Program of China under Grants 2019YFE0108300, the National Natural Science Foundation of China under Grants 52172379, 62001058 and U1864204, the Fundamental Research Funds for the Central University under Grants 300102242901.}
\thanks{Tao~Gao, Yuanbo~Wen and Ting~Chen are with the School of Information Engineering, Chang'an University, Xi'an 710064, China. E-mail: \{gtnwpu@126.com; wyb@chd.edu.cn; tchenchd@126.com\}}
\thanks{Kaihao~Zhang is with the School of Computing, Australian National University, Canberra, ACT, Australia. E-mail: \{kaihao.zhang@anu.edu.au\}}
\thanks{Peng~Cheng is with the Department of Computer Science and Information Technology, La Trobe University, Australia, and also 
 with the University of Sydney, Australia. E-mail: \{p.cheng@latrobe.edu.au; peng.cheng@sydney.edu.au\}}
}
\author{Tao Gao\orcidGT{},~\IEEEmembership{Member,~IEEE}, Yuanbo Wen\orcidWYB{}, Kaihao Zhang\orcidZKH{}, Peng Cheng\orcidCP{},~\IEEEmembership{Member,~IEEE}, and Ting Chen\orcidCT{}}



\maketitle

\begin{abstract}
Rain-by-snow weather removal is a specialized task in weather-degraded image restoration aiming to eliminate coexisting rain streaks and snow particles. In this paper, we propose RSFormer, an efficient and effective Transformer that addresses this challenge.
Initially, we explore the proximity of convolution networks (ConvNets) and vision Transformers (ViTs) in hierarchical architectures and experimentally find they perform approximately at intra-stage feature learning.
On this basis, we utilize a Transformer-like convolution block (TCB) that replaces the computationally expensive self-attention while preserving attention characteristics for adapting to input content.
We also demonstrate that cross-stage progression is critical for performance improvement, and propose a global-local self-attention sampling mechanism (GLASM) that down-/up-samples features while capturing both global and local dependencies. 
Finally, we synthesize two novel rain-by-snow datasets, RSCityScape and RS100K, to evaluate our proposed RSFormer.
Extensive experiments verify that RSFormer achieves the best trade-off between performance and time-consumption compared to other restoration methods.
For instance, it outperforms Restormer with a 1.53\% reduction in the number of parameters and a 15.6\% reduction in inference time.
Datasets, source code and pre-trained models are available at \url{https://github.com/chdwyb/RSFormer}.

\end{abstract}

\section{Introduction}
\IEEEPARstart{W}{eather} degradations, such as rain streaks and snow particles, severely affect the quality of image captured outdoors, which in turn limit the performance of subsequent advanced vision algorithms \cite{liu2018desnownet, fu2017clearing, che2019object, ding2016single, quan2023image}.
Due to the lack of specialized datasets and considerations of this special weather condition that rain and snow arise simultaneously, existing work \cite{fu2017clearing, ren2019progressive, chen2021all, chen2022snowformer, cui2022semi, jiang2020decomposition} eliminates the two types of weather separately.
In addition, the complex composition \cite{sun2010removal, zheng2013single, cai2021joint} of rain, snow and background makes it difficult to separate and further eliminate the degradations.
To the best of our knowledge, there is currently no such consideration for rain-by-snow weather removal.
In fact, a great deal of work \cite{cohen2015trends, wang2017hierarchical, barnum2007spatio, dai2008temperature, rainsnow, zheng2013single} and extensive experiments indicate that rain-by-snow weather removal is both practical and challenging, calling for an efficient and effective solution.

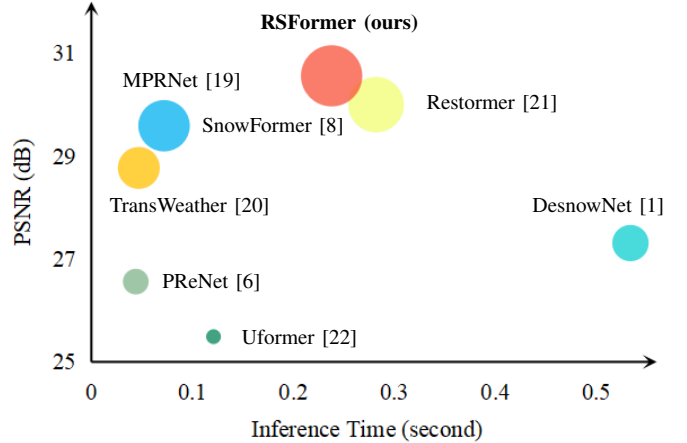
\begin{figure}[!t]
    \centering
    \hspace{-55mm}
    \begin{picture}(100,170)
    \put(0,0){\includegraphics[width=\linewidth]{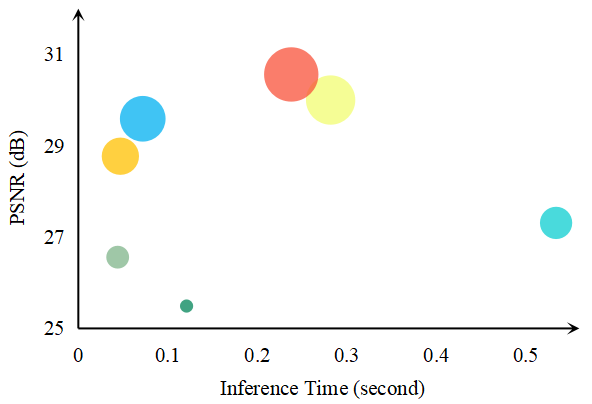}}
    \put(45,136){\footnotesize{MPRNet\cite{zamir2021multi}}}
    \put(200,89){\footnotesize{DesnowNet\cite{liu2018desnownet}}}
    \put(60,60){\footnotesize{PReNet\cite{ren2019progressive}}}
    \put(40,89){\footnotesize{TransWeather\cite{valanarasu2022transweather}}}
    \put(75,119){\footnotesize{SnowFormer\cite{chen2022snowformer}}}
    \put(160,128){\footnotesize{Restormer\cite{zamir2022restormer}}}
    \put(90,39){\footnotesize{Uformer\cite{wang2022uformer}}}
    \put(97,158){\footnotesize{\textbf{RSFormer (ours)}}}
    \end{picture}
\vspace{-3mm}
\caption{Efficiency comparisons of different state-of-the-art image restoration methods trained on our synthetic RS100K dataset while tested on RS100k-L test dataset, where our RSFormer achieves the best trade-off between inference time and peak signal noise ratio (PSNR).}
\label{fig:efficiency}
\end{figure}

Traditional methods \cite{zheng2013single, sun2010removal, xu2012removing, yu2014content} only work on some approximate scenes, failing to handle challenging restoration tasks due to the limited representation capability \cite{jiang2020multi}.
Recently, ConvNets utilize local receptive field and weight sharing mechanism to extract features and achieve reasonable success in both single image deraining and desnowing.
For instance, DerainNet \cite{fu2017clearing} separates the low-/high-frequency components to remove high-frequency rain streaks.
DesnowNet \cite{liu2018desnownet} adopts translucency and residual learning to recover image details degraded by snow particles and flakes.
However, local receptive field fails to model long-range dependencies and weight sharing cannot adapt to feature content.
To this end, ViTs \cite{vaswani2017attention} calculate the response of given pixel based on the global information of input features by self-attention (SA) mechanism.

Although ViTs have shown excellent performance on several up-stream vision tasks \cite{liu2021swin, carion2020end, dosovitskiy2020image, ding2022davit, dong2022cswin}, its resolution-squared computational complexity makes it infeasible for rain-by-snow weather removal with large resolution.
To handle this problem, extensive work focuses on improving the self-attention mechanism to reduce computation.
For instance, Swin Transformer \cite{liu2021swin} divides feature plains into small non-overlapping windows and then represents self-attention in separated windows.
Recently, ViTs gradually become the workhorse for image deraining and desnowing.
Among them, Uformer \cite{wang2022uformer} utilizes Swin Transformer and depth-wise multi-layer perceptron (MLP) to improve performance on several restoration tasks.
SnowFormer \cite{chen2022snowformer} explores the cross-attention to interact global-local context across patches.

Most existing Transformer-based image restoration methods \cite{wang2022uformer, zamir2022restormer, chen2022snowformer, liang2021swinir, valanarasu2022transweather} utilize self-attention to en-/de-code features.
However, recent work \cite{liu2022more, liu2022convnet, hou2022conv2former, yu2022metaformer} confirms that ConvNets designed following ViTs' architecture achieve comparable or better performance.
Here, we define:
1) Globality-wise global information means that the attention response of one given pixel is weighted by all other pixels.
2) Locality-wise global information means that the attention response of one given pixel weighted by several neighbor pixels. For instance, convolution \cite{liu2022convnet} and window-wise self-attention \cite{liu2021swin} are two ways to capture locality-wise global information.
According to extensive current work, we find that locality-wise global information is more significant than globality-wise global information in intra-stage feature learning.
In a different way, we imitate self-attention mechanism by the element-wise product between the output of a large convolution kernel and the value projection.
Furthermore, we introduce a Transformer-like convolution block (TCB).
Corresponding to SA and MLP in ViTs, TCB contains two main components, convolution attention mechanism (CAM) and convolution feed-forward network (CFFN).
The difference is that the attention weights of CAM are constant while self-attention weights change dynamically with  the input content.

With robust representation \cite{chen2022simple}, most Transformer-based methods follow a u-shaped design, which down-/up-sample feature maps with (transposed) convolution \cite{valanarasu2022transweather, chen2022snowformer} or pixel-(un)shuffle \cite{zamir2022restormer}. Since ConvNets' ability for intra-stage feature learning is consistent with ViTs \cite{liu2022more, liu2022convnet, hou2022conv2former, yu2022metaformer}, we indicate that accurate down-/up-sampling plays a key role in improving performance.
On this basis, we propose a global-local attention sampling mechanism (GLASM), which contains global-local attention down-sample (GLAD) and global-local attention up-sample (GLAU).
Different from \cite{valanarasu2022transweather, li2022rethinking}, we only down-/up-sample the value representation $V$ in self-attention and adopt light transposed self-attention (TSA) to efficiently capture long-range dependencies.
Furthermore, \cite{bai2022improving} demonstrates that self-attention is less effective in learning high-frequency information than ConvNets.
As a result, we incorporate a convolution layer that is optimized for high-frequency feature representation to address the limitations of ViTs.
By incorporating convolution, we can also recover the local information that is previously missing in self-attention. This allows our GLASM to effectively down-/up-sample accurate and information-rich features.

\begin{figure*}[!t]
\centering
\includegraphics[width=\linewidth]{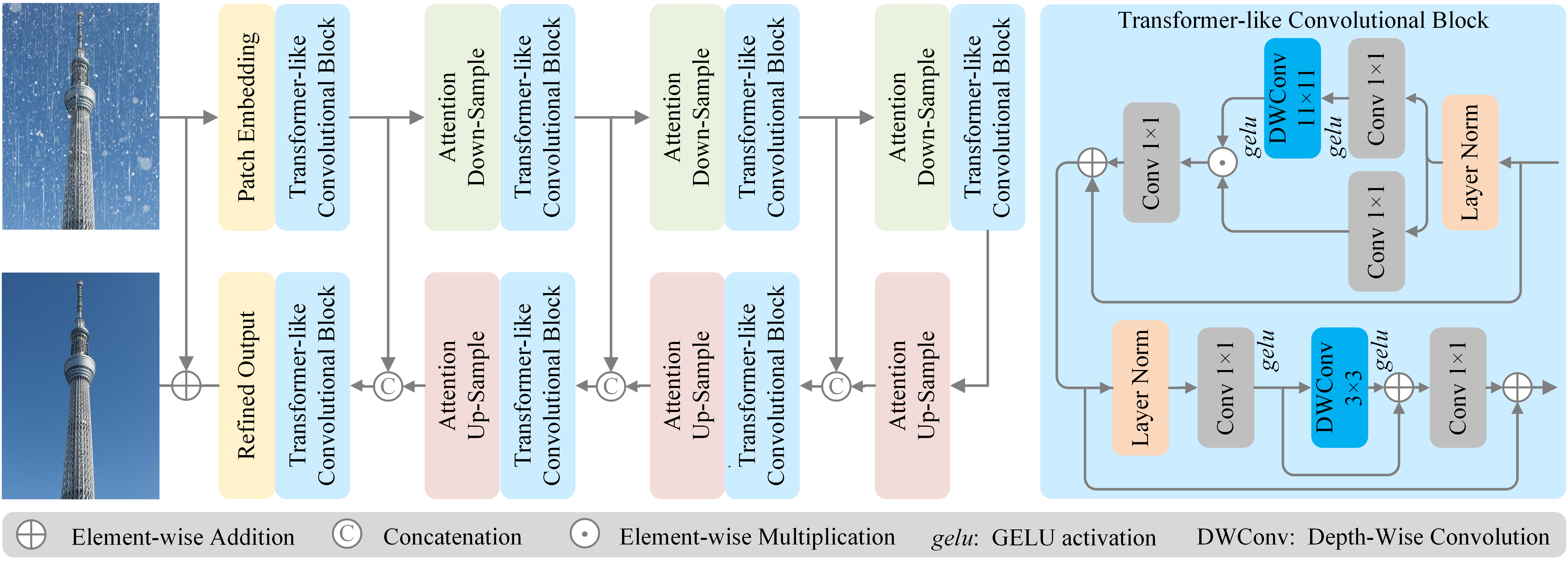}
\caption{Overview of our proposed effective and efficient RSFormer for rain-by-snow weather removal. Our RSFormer is designed for removing both rain streaks and snow particles from a single degraded image, which consists of two main components: 1) Transformer-like convolution block (TCB) that outperforms recent ViT backbones for feature learning at the same resolution level, 2) global-local self-attention sampling mechanism (GLASM) that first down-/up-samples feature maps with self-attention in U-shaped image restoration methods.}
\label{fig:network}
\end{figure*}

Fig. \ref{fig:network} illustrates the overview of our proposed RSFormer for restoring the image captured under rain-by-snow weather condition. In this paper, our main contributions are summarized as follows.

\vspace{2mm}
\begin{itemize}
\item We initially focus on rain-by-snow weather-degraded image restoration and propose an effective and efficient Transformer to restore the degraded image, named RSFormer, which achieves the best trade-off between the quality of output image and the time required for inference comparing with existing methods.
\item Based on extensive current work, we conclude that ConvNets and ViTs perform approximately on intra-stage feature learning in hierarchical architectures. Meanwhile, we demonstrate that cross-stage progression is crucial for performance improvement. To achieve fine down-/up-sampling while preserving information-rich features, we develop a global-local attention sampling mechanism (GLASM).
\item Two novel rain-by-snow datasets, RSCityScape and RS100K, are synthesized for rain-by-snow weather removal. Additionally, we collect a real-world dataset for practical application evaluation.
\end{itemize}
\vspace{2mm}

 \section{Related Work}
 \subsection{Image Deraining}
Recently, deep learning has been overwhelmingly successful in image restoration \cite{zhang2021benchmarking,zhang2022edface,zhang2022deep,zhang2020deblurring}, which also includes image deraining \cite{deng2019drd,jiang2020multi,fu2017clearing,yang2017deep,yasarla2019uncertainty,zhang2021dual,ren2019progressive,zhang2020beyond}.
DerainNet \cite{fu2017clearing} and JORDER \cite{yang2017deep} are two of the earliest convolution-based methods for deep single image deraining.
To improve generalization, SEMI \cite{wei2019semi} exploits both synthetic and real-world rainy images to conduct supervised and unsupervised training respectively.
In addition, rain presents notable diversity (\ie 
density, size, distribution, \etc.), thus prior learning may be helpful for accurate rain removal.
 Among these methods, DIDMDN \cite{zhang2018density} guides the network to restore degraded image by estimating rain density.
 Furthermore, UMRL \cite{yasarla2019uncertainty} and MSPFN \cite{jiang2020multi} utilize uncertainty and multi-scale information to obtain derained images.
Subsequently, the squeeze-excitation mechanism and progressive recursive learning are introduced in RESCAN \cite{li2018recurrent}, PReNet \cite{ren2019progressive} and DPENet \cite{gao2022heavy} to design networks.
MSPFN \cite{jiang2020multi} introduces multi-scale information to adapt to the distribution diversity of rain streaks.
Next, multi-stage architectures are further used to improve overall performance.
Specifically, DRDNet \cite{deng2019drd} adopts two branches to remove rain streaks and recover image details.
MPRNet \cite{zamir2021multi} establishes a multi-stage progressive restoration network and specializes in a high-quality reconstruction network. 
Recent work, MAXIM \cite{tu2022maxim}, also follows the same multi-stage design with multi-layer perceptron to remove degradations.
Lately, ViTs gradually replace ConvNets as the workhorse for image deraining.
For instance, based on Swin Transformer \cite{liu2021swin}, Uformer \cite{wang2022uformer} achieves fine performance on several tasks including rain removal.
By exploring the inner product nature of self-attention, Restormer \cite{zamir2022restormer} leverages transposed self-attention to build an efficient restoration algorithm.

\subsection{Image Desnowing}
Snow removal is also a classic image restoration task.
DesnowNet \cite{liu2018desnownet} performs as one of the earliest deep learning methods shifting attention to desnowing, which is similarly based on multi-stage convolution network.
Meanwhile, it contributes the first desnowing benchmark, Snow100K dataset.
Subsequently, multi-scale features and transparency perception are exploited by JSTASR \cite{chen2020jstasr} to remove snow particles and snow flakes.
HDCWNet \cite{chen2021all} proposes a hierarchical desnowing network based on dual-tree discrete wavelet transform, and further establishes a high-/low-frequency reconstruction module to generate high-quality desnowed image.
Next, DDMSNet \cite{zhang2021deep} considers semantic features and depth information to restore snowy images with advanced visual features.
Furthermore, it also demonstrates that several deraining methods (RESCAN \cite{li2018recurrent} and SPANet \cite{wang2019spatial}) achieve competitive performance in desnowing community.
Recently, SMGARN \cite{cheng2022snow} utilizes snow masks to locate and remove snow.
Based on cross-attention, SnowFormer \cite{chen2022snowformer} develops an efficient desnowing Transformer-based network.
In addition, All-in-One \cite{li2020all}, TKL \cite{chen2022learning} and TransWeather \cite{valanarasu2022transweather} focus on multiple weather-degraded image restoration and perform well in single image desnowing.

\subsection{Vision Transformer}
ViTs \cite{han2022survey} have shown their potential for down-stream vision tasks due to the competitive ability to effectively model long-range dependencies, leading to several derived image restoration methods, like Uformer \cite{wang2022uformer}, Restormer \cite{zamir2022restormer}, TransWeather \cite{valanarasu2022transweather} and SnowFormer \cite{chen2022snowformer}.
Several recent work \cite{hou2022conv2former, yu2022metaformer, liu2022convnet} demonstrates that ConvNets may achieve comparable or even surpassing performance by clever design based on ViTs architecture.
For example, ConvNeXt \cite{liu2022convnet} replaces components in the residual network following ViTs construction and designs network while Conv2Former \cite{hou2022conv2former} builds a Transformer-like convolution module and performs well on several high-level vision tasks.
However, Transformer-like ConvNets \cite{hou2022conv2former}, token-mixer exploration \cite{yu2022metaformer, yu2022metaformer, wang2022shift} and components replacement \cite{liu2022convnet, liu2022more} keep learning at the intra-stage resolution, and fail to concentrate on cross-stage progression, namely down-/up-sampling.

 \section{Proposed Method}
 \label{sec:method}
The overview of our RSFormer is shown in Fig. \ref{fig:network}.
In this section, we describe the architectures of TCB, GLASM, and the spatial-frequency loss function used for training.

 \subsection{Transformer-like Convolution Block}
 The success of ViTs \cite{zamir2022restormer, wang2022uformer, valanarasu2022transweather, chen2022snowformer} in image restoration is experimentally attributed to the robust representation for modeling long-range dependencies \cite{d2021convit}.
 However, several studies demonstrate that ConvNets may achieve comparable or even superior performance by following a similar construction as Transformer, like ConvNeXt \cite{liu2022convnet} and Conv2Former \cite{hou2022conv2former}. 
 Inspired by these work, we find that ConvNets modified subtly perform well on intra-stage feature learning in hierarchical structures. 
 To be sure, convolution is more memory-friendly than calculation-heavy self-attention mechanism, specifically at large resolution feature for certain.
 Therefore, we utilize a  Transformer-like convolution block following \cite{hou2022conv2former}, which (seen in Fig. \ref{fig:network}) mainly contains two components: convolution attention module (CAM) and convolution feed-forward network (CFFN).
 The query representa $Q$ and key matrix $K$ in self-attention mechanism are to generate an attention diagram for value representation $V$, which can be given by
 \begin{equation}
 \label{eq:sa}
     SA(Q, K, V) = Q\otimes K^T\otimes V,
 \end{equation}
 where $\otimes$ denotes matrix multiplication, and $T$ is the matrix transposing operation.
 The successful experience of splitting windows in Swin Transformer \cite{liu2021swin} to represent self-attention indicates that it is unnecessary to capture globality-wise global but locality-wise global information.
 Following \cite{yu2022metaformer, hou2022conv2former}, we simply reduce $Q$ and $K$ to single-source attention $A$, which presents attention maps through depth-wise separable convolution with large kernel size and can be formulated as
 \begin{equation}
 \label{eq:ca}
     CA(A, V)=A\odot V,
 \end{equation}
 where CA and $\odot$ denote convolution attention and element-wise product, respectively. It specifically means that the response of given pixel is based on the weighted sum of pixels in one square area included by convolution kernel instead of all pixels. 
It is clear that our design is a representation of locality-wise global information and can be supported by \cite{liu2021swin} to a certain extent.
 Structurally, CAM maintains the same input/output projection design to original self-attention.
 Compared to Swin Transformer \cite{liu2021swin} and ConvNeXt \cite{liu2022convnet}, our approach is significantly memory-friendly and keeps the attention property that adapts to input features.
 In the feed-forward network, we utilize depth-wise convolution to fuse spatial information while original attention only merges channel features.
 CFFN is also a common design at present.
 Our proposed TCB can be formulated as
 \begin{align}
     & X^{'} = X^{l-1} + CA(LN(X^{l-1})) \times \gamma_1, \\
     & X^l = X^{'} + FFN(LN(X^{'})) \times \gamma_2,
 \end{align}
 where $X^{l-1}$, $X^{'}$, and $X^{l}$ denote the input, mid-output, and output features of the TCB, respectively, $LN$ indicates layer normalization, and $\gamma$ is layer scale \cite{touvron2021going}.
 The above two components form the TCB, and different stages share similar structures as Transformer-based methods \cite{zamir2022restormer, wang2022uformer} for our target rain-by-snow weather removal.

 \subsection{Global-Local Self-Attention Sampling Mechanism}

\begin{figure*}[t]
\centering
\subfloat[Global-Local self-Attention Down-sample (GLAD)]{
	\begin{minipage}[b]{0.5\linewidth}
		\includegraphics[width=3.5in]{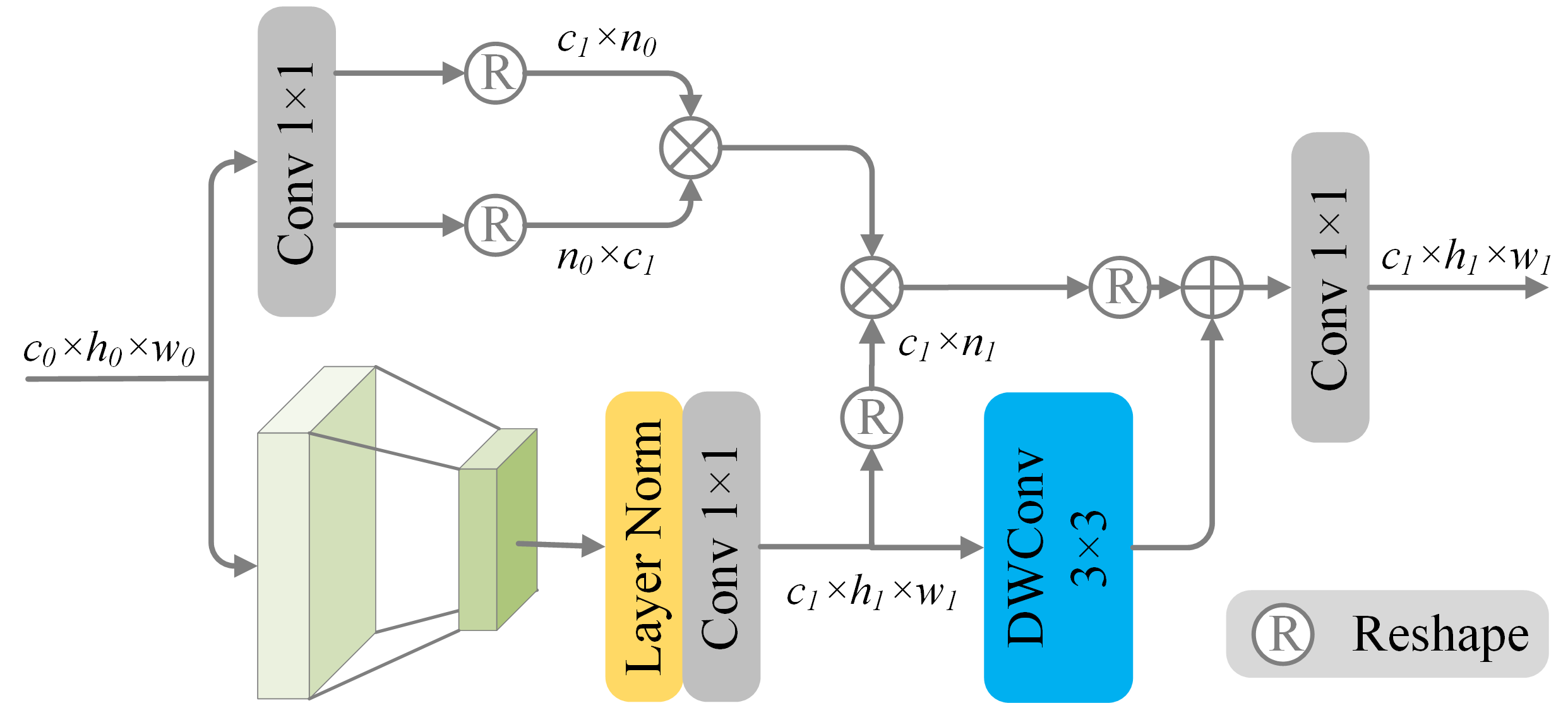}
	\end{minipage}}
\subfloat[Global-Local self-Attention Up-sample (GLAU)]{
	\begin{minipage}[b]{0.5\linewidth}
		\includegraphics[width=3.5in]{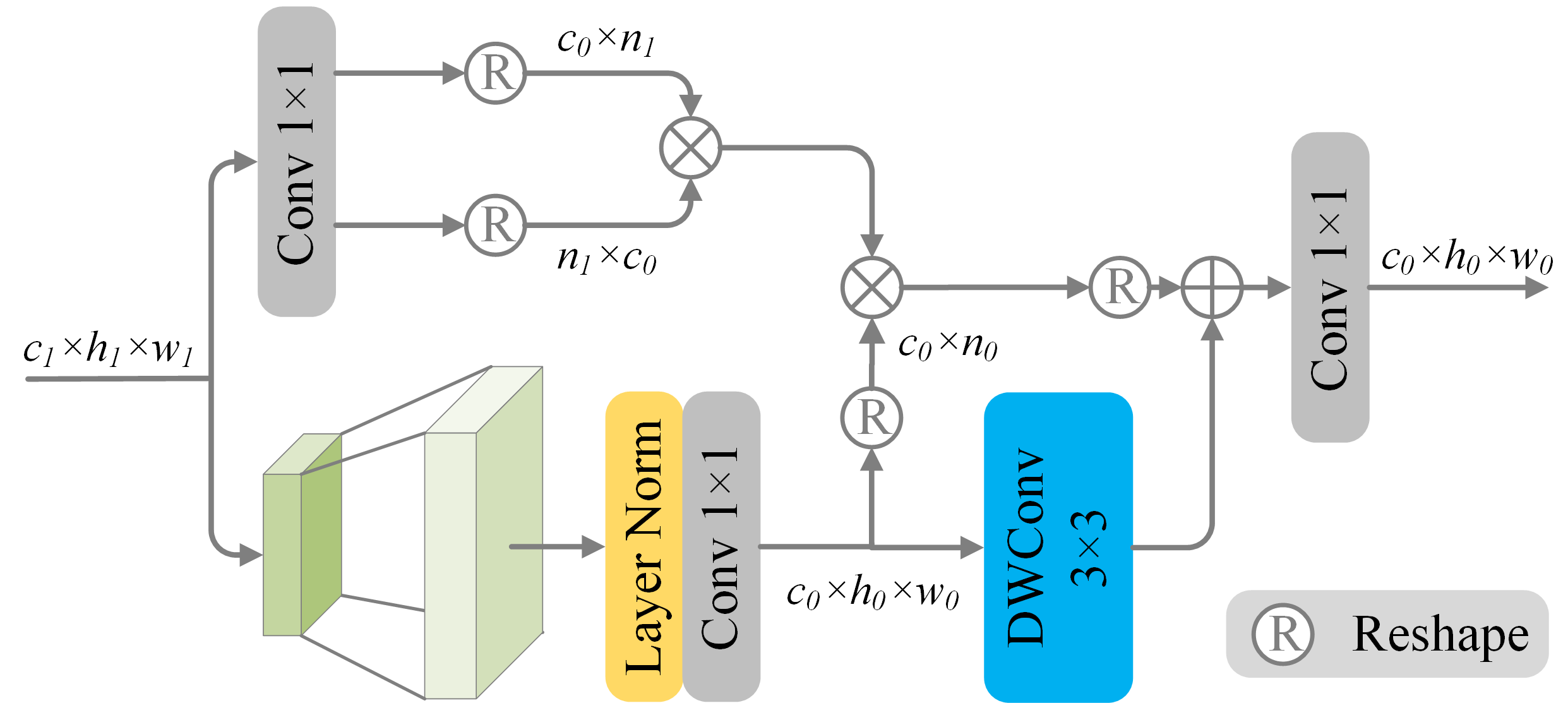}
	\end{minipage}}
\caption{Illustrations of our proposed global-local self-attention sampling mechanism (GLASM), which contains two parts, down-sampling method GLAD and up-sampling method GLAU. Our GLASM follows the same changes of feature size as usual methods \cite{wang2022uformer, valanarasu2022transweather, zamir2022restormer}. Specifically in this figure, $n_1=4\times n_0$, $n_0=h_0\times w_0$, $n_1=h_1\times w_1$, $c_1=2\times c_0$.}
\label{fig:sample}
\end{figure*}
 
Existing work \cite{yu2022metaformer, liu2021swin, hou2022conv2former} on ViTs focus on intra-stage feature learning of hierarchical stages, ignoring the exploration of cross-stage progression.
The common operators used for cross-stage progression are (transposed) convolution \cite{valanarasu2022transweather, wang2022uformer} or pixel-(un)shuffle \cite{zamir2022restormer}.
Although \cite{li2022efficientformer} shifts attention here but also conducts down-/up-sampling without multi-scale information.
In fact, accurate and appropriate sampling also plays a significant role in improving performance.
To this end, we propose a global-local self-attention sampling mechanism (GLASM) as Fig. \ref{fig:sample} shown.

In particular, we find that the inner product in transposed (channel) self-attention (TSA) mechanism allows computation from different resolution-level. In this case, we represent value projection $V$ as the target resolution while $Q$ and $K$ are from lower-/upper-level features.
We also find that transposed self-attention is computation-friendly.
The comparison of transposed self-attention and spatial self-attention (SSA) on computational complexity can be found in TABLE \ref{tab:computation}, where the size of input feature is (1, 64$\times64$, 32). Obviously, transposed self-attention is lighter than spatial self-attention, thereby we adopt the former to calculate attention-enhanced features. Therefore, we represent self-attention along channel dimension.
The comparison of these two attention mechanisms on performance is shown in Sec. \ref{sec:components}.
\begin{table}[H]
     \centering
     \caption{Comparison of computational amount of transposed self-attention and spatial self-attention. The latter is severely computation-heavy over the transposed self-attention.}
     \label{tab:computation}
     \renewcommand\arraystretch{1.25}
     \setlength{\tabcolsep}{5mm}{
     \begin{tabular}{lcc}
     \toprule
     Self-attention & Production & Addition \\
    \midrule
    Transposed & $8.39\times10^6$ & $8.26\times10^6$ \\
    Spatial & $1.07\times10^9$ & $1.06\times10^9$ \\
    \bottomrule
    \end{tabular}
     }
 \end{table}
Besides, several work \cite{li2022efficientformer, yang2021focal, zhao2021hybrid} indicate that self-attention alone for enhancing features results in a lack of local dependencies, which makes it impossible to model high-frequency details of images\cite{bai2022improving}.
Therefore, we utilize convolution to supplement value feature $V$ to ensure that the sampling operation achieves both global and local feature-rich cross-stage progression.
We formulate our down-sampling method (GLAD) as
\begin{equation}
    down(Q_{n_0, c_1}, K_{n_0, c_1}, V_{n_1, c_1}) = Q^{T}\otimes K\otimes V^{T} + L(V),
\end{equation}
where $n_0$, $c_0$ denote the sizes of low-resolution features, $n_1$, $c_1$ indicate the sizes of up-resolution features, $L(\cdot)$ is the operator for capturing local information implemented by depth-wise convolution.
Obviously, our GLASM is a kind of cross-scale self-attention, which effectively exploits the features with different scales to achieve accurate sampling and further improve the performance for rain-by-snow weather removal.
Our up-sampling method GLAU follows a similar operation as down-sampling but value representation $V$ presents as high-resolution feature maps.
GLAU is formulated as
\begin{equation}
    up(Q_{n_1, c_0}, K_{n_1, c_0}, V_{n_0, c_0}) = Q^{T}\otimes K\otimes V^{T} + L(V).
\end{equation}
Experimental results confirm that our TCB achieves similar performance and fast inference to self-attention-based architectures for intra-stage feature learning while our GLASM outperforms the common-used down-/up-sampling operation due to its robustness with both global and local information.

 \subsection{Spatial-Frequency Loss}
\begin{figure}[H]
\centering
    \subfloat[Degraded]{\includegraphics[height=1.2in]{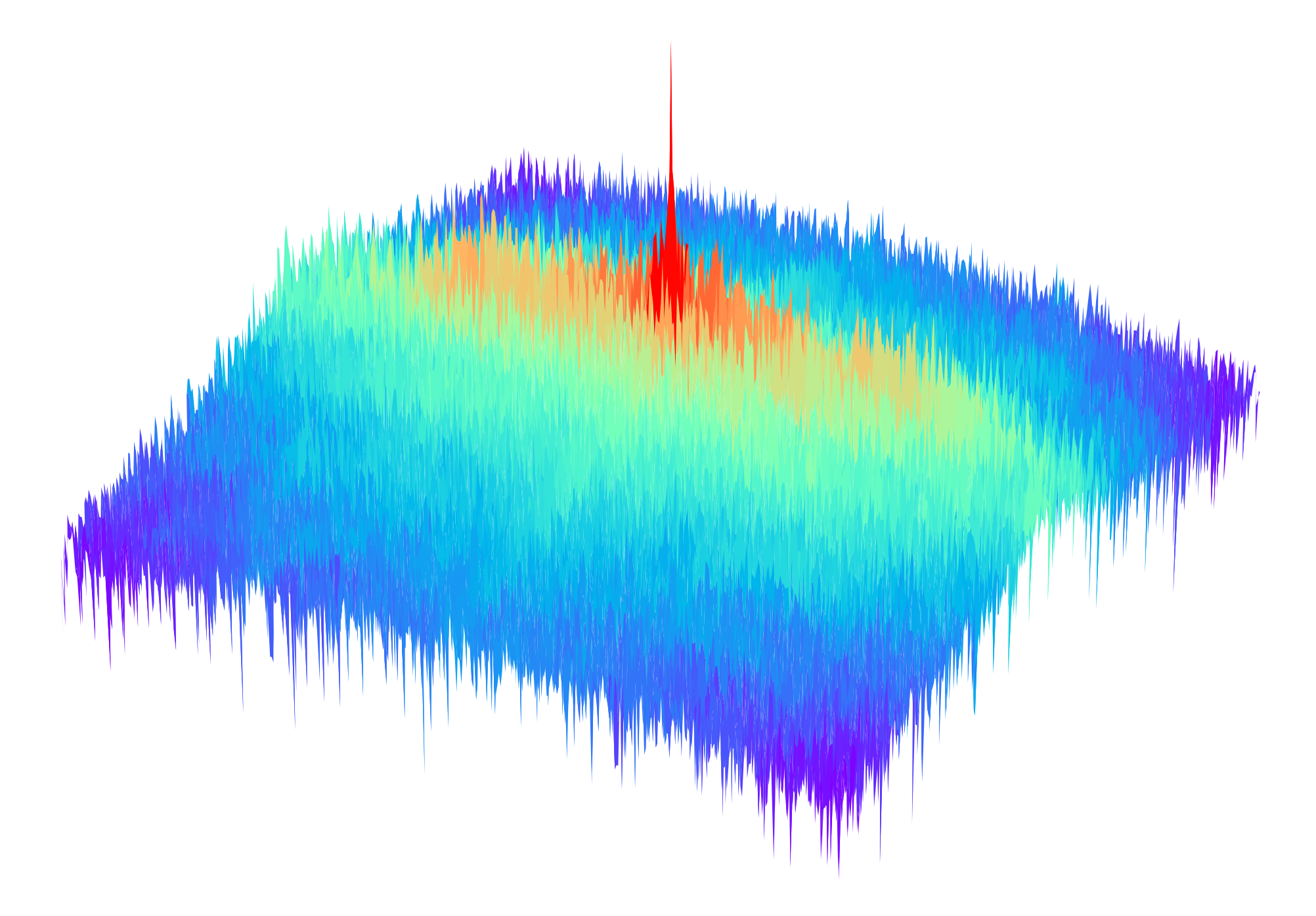}}
    \subfloat[Degradation-free]{\includegraphics[height=1.2in]{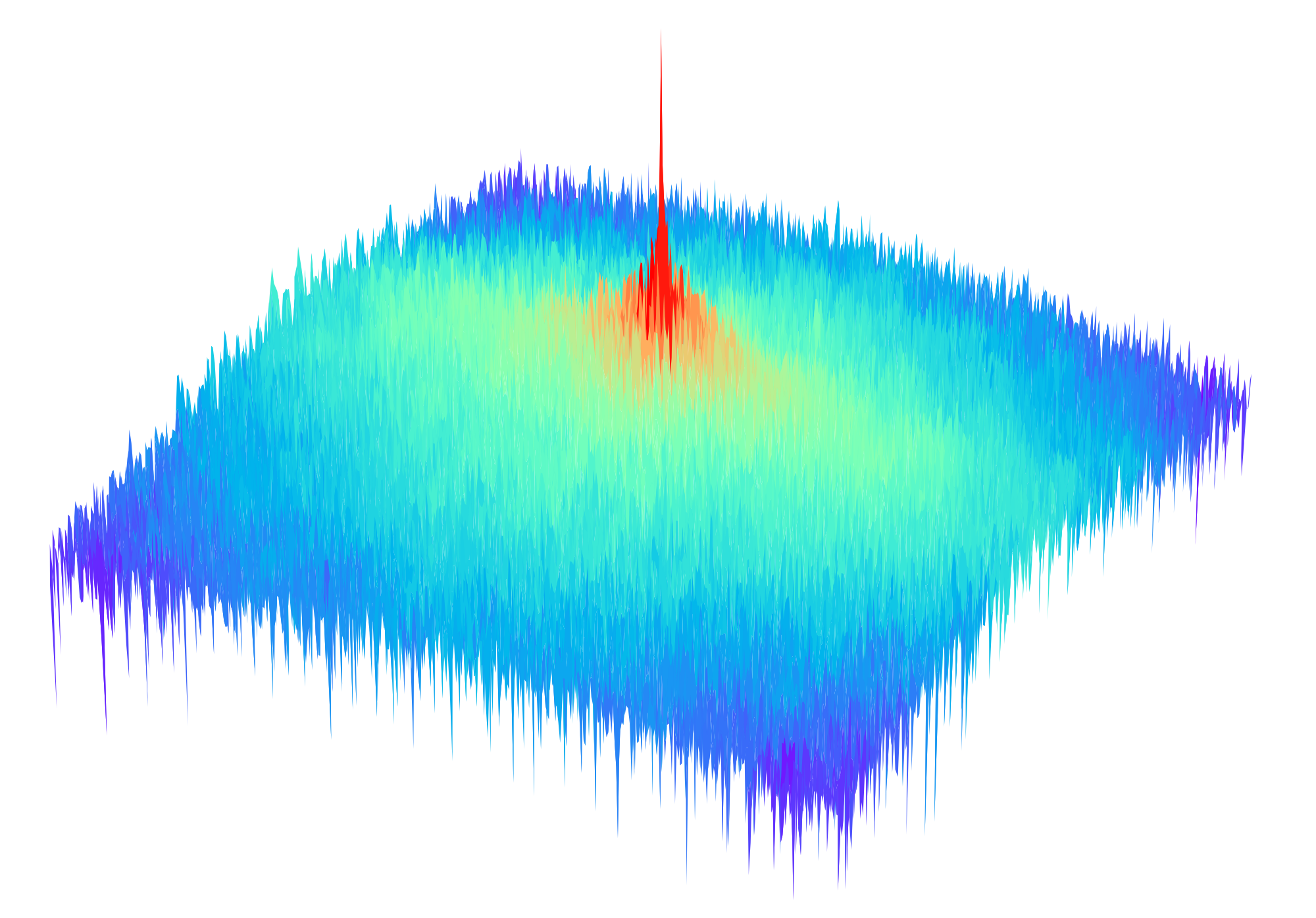}}
    \caption{Spectrums of rain-by-snow image before and after removing degradations, which illustrates the difference and suggests that spectral similarity is also an significant criterion for image restoration.}
    \label{fig:spectrum}
\end{figure}

Rain streaks and snow particles present significant marginality \cite{jiang2021focal}, which is reflected in the spectral difference before and after restoration (seen in Fig. \ref{fig:spectrum}).
However, commonly used loss functions for image restoration measure the similarity between the restored and degradation-free image in the spatial domain.
The apparent spectral variation is not taken into account.
Therefore, we propose a spatial-frequency loss function, which constrains the discrepancy in both spatial and frequency domains.
Specifically, we utilize focal frequency loss \cite{jiang2021focal} in the frequency domain.
The spectrum distance of restored image and ground truth can be formulated as
\begin{equation}
    d(F_r, F_g)=\frac{1}{MN}\sum_{u=0}^{M-1}\sum_{v=0}^{N-1}\left|F_r(u, v)-F_g(u, v) \right|^2,
\end{equation}
where $F_r$ and $F_g$ denote the spectrum of the restored image and corresponding ground truth, respectively. $M$ and $N$ are the sizes of the spectrum.
Meanwhile, dynamic weight is utilized to focus on significant frequency, which is defined as
\begin{equation}
    w(u, v)=\left|F_r(u, v)-F_g(u, v) \right|^{\alpha},
\end{equation}
where $\alpha$ is the scaling factor for flexibility.
Then we have the focal frequency loss
\begin{equation}
    \mathcal{L}_{freq} = w(u, v)\cdot d(F_r, F_g).
\end{equation}

We use Charbonnier loss \cite{charbonnier1994two} as the loss function in the spatial domain
\begin{equation}
    \mathcal{L}_{spat}=\sqrt{\left||I_r-I_g\right||^2+\epsilon^2},
\end{equation}
where $I_r$ and $I_g$ are the images in spatial domain, and $\epsilon$ is empirically set to $10^{-3}$ for all experiments \cite{zamir2021multi}.
Therefore, our spatial-frequency loss function is
\begin{equation}
    \mathcal{L}=\mathcal{L}_{spat} + \beta\cdot \mathcal{L}_{freq},
\end{equation}
where $\beta$ is the loss weight of focal frequency loss.
Our spatial-frequency loss function ensures the similarity of two domains simultaneously, thus facilitating the generation of high-quality restored images.

 \section{Experimental Results}
 \subsection{Implementation Specifications}
We train our proposed RSFormer for rain-by-snow weather removal on 128$\times$128 random-cropped patches with RandAugment \cite{cubuk2020randaugment} and batch size of 32.
The Adam \cite{kingma2014adam} optimizer with an initial learning rate of $2\times 10^{-4}$ is used to optimize parameters, where the learning rate is gradually decreased to $1\times 10^{-7}$ by cosine annealing decay \cite{loshchilov2016sgdr}.
All methods are trained for a total of 200 epochs with their own configurations. 
In inference, we pad and un-pad degraded and restored images respectively for adapting to all resolutions.
The training is performed on NVIDIA Tesla A100, while the inference is conducted on NVIDIA RTX 3090.
 
 \subsection{Datasets}
To train and evaluate the performance of  RSFormer, we synthesize two rain-by-snow datasets, referred to as RSCityScape and RS100K.
Meanwhile, we also collect a real-world rain-by-snow dataset to evaluate our RSFormer for real applications.

\subsubsection{RSCityScape} Our synthentic rain-by-snow dataset RSCityScape is based on RainCityScape \cite{hu2019depth} dataset and the snowy approach from DesnowNet \cite{liu2018desnownet}. 
RainCityScape dataset consists of outdoor images with rain and fog, which contains 9432 and 1188 rain-by-snow image pairs for training and testing, respectively.
We randomly select snow masks from the training dataset of Snow100K and add them to rainy images, further generating our RSCityScape dataset for weather removal where rain streaks and snow particles appear simultaneously.
Meanwhile, we reduce image resolution to a quarter of the original due to practical limitations.

\subsubsection{RS100K} Snow100K \cite{liu2018desnownet} dataset is the earliest benchmark for single image desnowing. With the strategy of generating rain streaks in \cite{guo2021efficientderain}, we synthesize the other rain-by-snow dataset RS100K. It contains 50000 image pairs to train the proposed RSFormer while three testing datasets to evaluate performance named RS100K-L, RS100K-M, and RS100K-S, respectively.
All the testing datasets include 2000 image pairs randomly selected from the original quantity.

\subsubsection{Real-World RS300} To evaluate our RSFormer on real-world images with both rain streaks and snow particles, we collect a real-world RS300 dataset from Baidu and Google explorer.
The scene of image acquisition includes night, traffic, street, pedestrian and other applications where rain and snow coexist in reality.

 
 \subsection{Rain-by-snow Weather Removal}
Quantitative comparisons are performed by utilizing peak signal noise ratio (PSNR) and structural similarity (SSIM) metrics.
We evaluate the proposed RSFormer and other methods with state-of-the-art performance on deraining/desnowing.
Among the compared methods, DesnowNet \cite{liu2018desnownet}, TransWeather \cite{valanarasu2022transweather} and SnowFormer \cite{chen2022snowformer} are related to single image desnowing, while PReNet \cite{ren2019progressive}, MPRNet \cite{zamir2021multi}, Uformer \cite{wang2022uformer} and Restormer \cite{zamir2022restormer} perform well on rain streaks removal.
TABLE \ref{tab:synthetic} reports the PSNR/SSIM scores on our synthetic RSCityScape and RS100K datasets.
On RSCityScape dataset, our RSFormer achieves considerable gains of 5.20 dB and 2.51 dB over SnowFormer \cite{chen2022snowformer} and Restormer \cite{zamir2022restormer}, respectively.
Meanwhile, RSFormer also achieves the best quantitative performance on all three testing datasets of RS100K.
To sum up, our proposed RSFormer obtains performance gains of $3.53\%\sim 12.2\%$ and $0.52\%\sim 2.98\%$ over the other compared methods in PSNR/SSIM.
Furthermore, we present visual comparisons in Fig. \ref{fig:rscityscape} and Fig. \ref{fig:rs100k}.
Fig. \ref{fig:rscityscape} illustrates that all methods can remove both rain/fog and snow from traffic scenes; however, our RSFormer preserves rich details and better fidelity while other methods are confused between background and degradations.
In particular, road signs in the second column are missing after rain-by-snow removal by other methods while our RSFormer protects it well.
In Fig. \ref{fig:rs100k}, compared to others, our RSFormer does well in eliminating dark spots after snow particles removal (seen in the 1st, 3rd and 5th columns), as well as removing hard rain streaks (seen in the 6th and 7th columns).
To further observe the advantages of our approach, we have also displayed the corresponding error maps of Fig. \ref{fig:rscityscape} and Fig. \ref{fig:rs100k} on our github homepage.
Overall, the image generated by our proposed RSFormer preserve better details and are closer to the corresponding ground truth.

 \begin{table*}[!t]
     \caption{Rain-by-snow weather removal results. Best and second highest scores are \textbf{highlighted} and \underline{underlined}. For each method, reduction in error relative to the best-performing methods is reported in parenthesis. Our RSFormer achieves 3.52\% ~ 17.1\% relative improvement in PSNR over the other methods.\label{tab:synthetic}}
     \centering
     \renewcommand\arraystretch{1.25}
     \setlength{\tabcolsep}{2.2mm}{
     \begin{tabular}{lccccccccccc}
     \toprule
     \multirow{2}{*}{Methods} & \multirow{2}{*}{Venue} & \multicolumn{2}{c}{RSCityScape} & \multicolumn{2}{c}{RS100K-L} &  \multicolumn{2}{c}{RS100K-M} & \multicolumn{2}{c}{RS100K-S} & \multicolumn{2}{||c}{Average} \\
     ~  & ~ & PSNR & SSIM & PSNR & SSIM & PSNR & SSIM & PSNR & SSIM & \multicolumn{1}{||c}{PSNR (\ $\uparrow$\ )} & SSIM (\ $\uparrow$\ ) \\
    \midrule
    DesnowNet\cite{liu2018desnownet} & \textit{TIP 2018}  & 30.14 & 0.939 & 27.31 & 0.913 & 30.22 & 0.956 & 31.97 & 0.952 & \multicolumn{1}{||c}{29.91 \colorbox{gray!20}{(12.2\%)}} & 0.940 \colorbox{gray!20}{(2.98\%)}  \\
    PReNet \cite{ren2019progressive} & \textit{CVPR 2019} & 28.49 & 0.932 & 26.56 & 0.907 & 29.47 & 0.942 & 30.12 & 0.948 & \multicolumn{1}{||c}{28.67 \colorbox{gray!20}{(17.1\%)}} & 0.932 \colorbox{gray!20}{(3.86\%)} \\
    MPRNet\cite{zamir2021multi} & \textit{CVPR 2021}  & 31.03 & 0.969 & 	29.59 & 0.931	& 32.85 & 0.962 & 33.86 & 0.967 & \multicolumn{1}{||c}{31.83 \colorbox{gray!20}{(5.44\%)}} & 0.957 \colorbox{gray!20}{(1.15\%)} \\
    Uformer\cite{wang2022uformer} & \textit{CVPR 2022}  & 31.63 & 0.972 &	25.49 & 0.890 & 31.07 & 0.945 & 32.71 & 0.956 & \multicolumn{1}{||c}{30.23 \colorbox{gray!20}{(11.0\%)}} & 0.941 \colorbox{gray!20}{(2.87\%)} \\
    TransWeather\cite{valanarasu2022transweather} & \textit{CVPR 2022}  & 28.46 & 0.944 & 28.97 & 0.921 & 31.53 & 0.951 & 32.32 & 0.956 & \multicolumn{1}{||c}{30.32 \colorbox{gray!20}{(10.7\%)}} & 0.943 \colorbox{gray!20}{(2.65\%)} \\
    SnowFormer\cite{chen2022snowformer} & \textit{arXiV 2022}  & 29.34 & 0.960 & 29.10 & 0.932 & 32.37 & 0.960 & 33.49 & 0.966 & \multicolumn{1}{||c}{31.08 \colorbox{gray!20}{(7.98\%)}} & 0.955 \colorbox{gray!20}{(1.36\%)} \\
    Restormer\cite{zamir2022restormer} & \textit{CVPR 2022}  & \underline{32.03} & \underline{0.977}	& \underline{30.00} & \underline{0.940} & \underline{33.24} & \underline{0.965} & \underline{34.40} & \underline{0.970} & \multicolumn{1}{||c}{\underline{32.42} \colorbox{gray!20}{(3.52\%)}} & \underline{0.963} \colorbox{gray!20}{(0.52\%)} \\
    \textbf{RSFormer} & ---  & \textbf{34.54} & \textbf{0.981} & \textbf{30.56} & \textbf{0.946} & \textbf{33.97} & \textbf{0.969} & \textbf{35.18} & \textbf{0.974} & \multicolumn{1}{||c}{\textbf{33.56} \colorbox{gray!20}{(0.00\%)}} & \textbf{0.968} \colorbox{gray!20}{(0.00\%)} \\
    \bottomrule
    \end{tabular}
     }
 \end{table*}

 \begin{figure*}[!t]
     \includegraphics[height=0.7in]{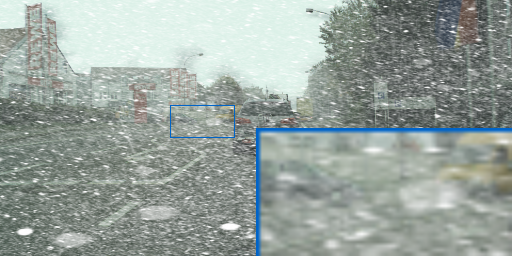}
     \includegraphics[height=0.7in]{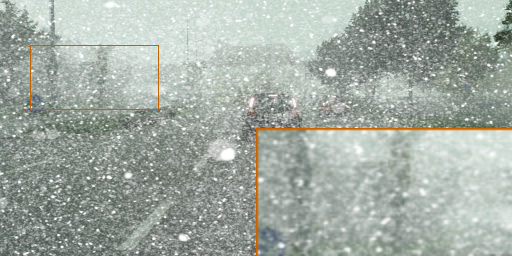}
     \includegraphics[height=0.7in]{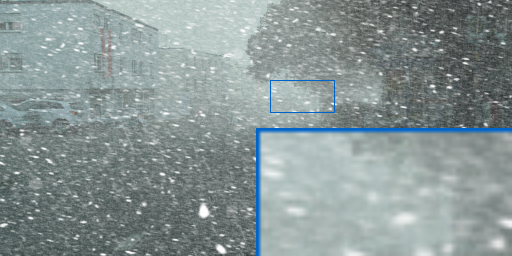}
     \includegraphics[height=0.7in]{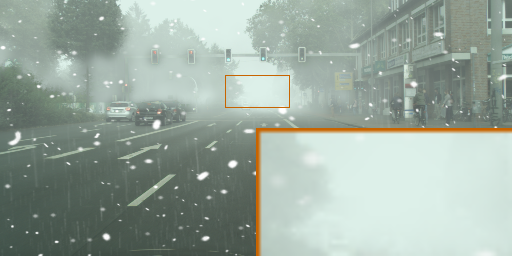}
     \includegraphics[height=0.7in]{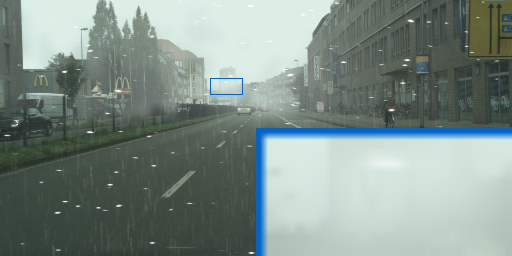}
     
     \vspace{-1.2mm}
     \renewcommand\arraystretch{0.5}
     \setlength{\tabcolsep}{7.8mm}{
     	\begin{tabular}{ccccc}
     		10.13 / 0.3990 & 9.031 / 0.3118 & 10.03 / 0.3756 & 10.57 / 0.6771 & 16.93 / 0.8584\\
     	\end{tabular}
     }\vspace{-4.8mm}

     \vspace{1mm}
     \includegraphics[height=0.7in]{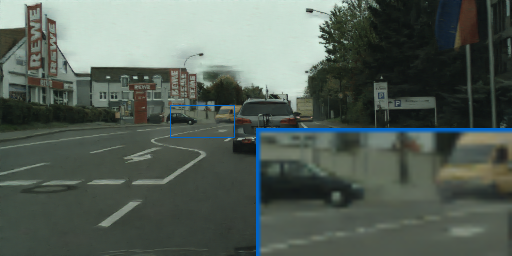}
     \includegraphics[height=0.7in]{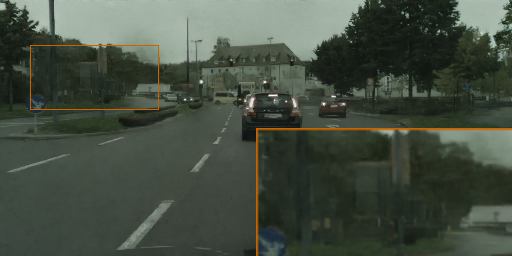}
     \includegraphics[height=0.7in]{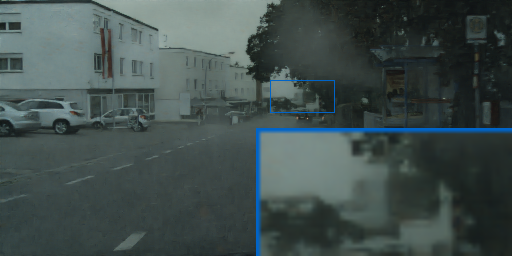}
     \includegraphics[height=0.7in]{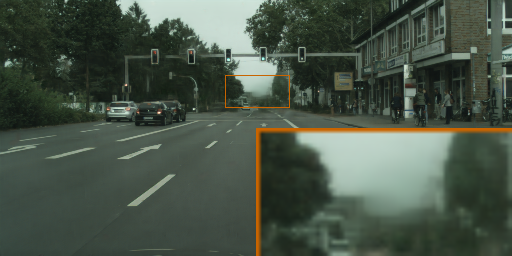}
     \includegraphics[height=0.7in]{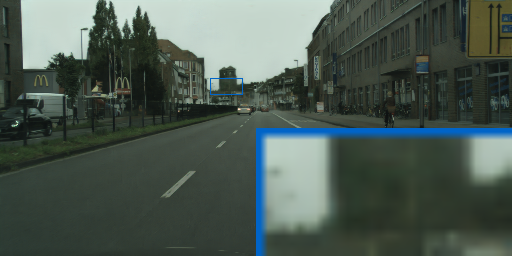}
     
     \vspace{-1.2mm}
     \renewcommand\arraystretch{0.5}
     \setlength{\tabcolsep}{7.8mm}{
     	\begin{tabular}{ccccc}
     		22.74 / 0.9398 & 27.40 / 0.9335 & 23.05 / 0.8916 & 30.39 / 0.9711 & 31.23 / 0.9831\\
     	\end{tabular}
     }\vspace{-4.8mm}

     \vspace{1mm}
     \includegraphics[height=0.7in]{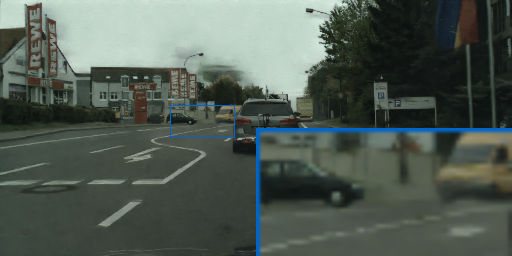}
     \includegraphics[height=0.7in]{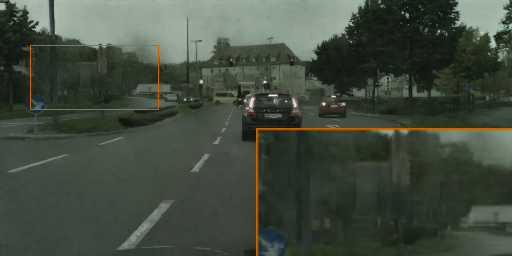}
     \includegraphics[height=0.7in]{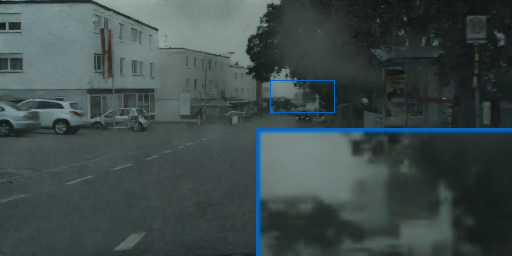}
     \includegraphics[height=0.7in]{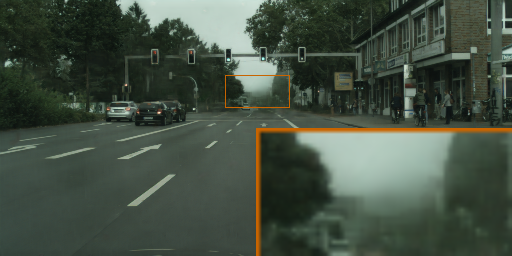}
     \includegraphics[height=0.7in]{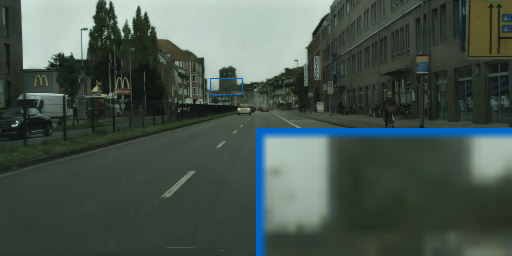}
     
     \vspace{-1.2mm}
     \renewcommand\arraystretch{0.5}
     \setlength{\tabcolsep}{7.8mm}{
     	\begin{tabular}{ccccc}
     		24.01 / 0.9396 & 25.66 / 0.9265 & 23.59 / 0.8920 & 27.97 / 0.9639 & 28.56 / 0.9800\\
     	\end{tabular}
     }\vspace{-4.8mm}

     \vspace{1mm}
     \includegraphics[height=0.7in]{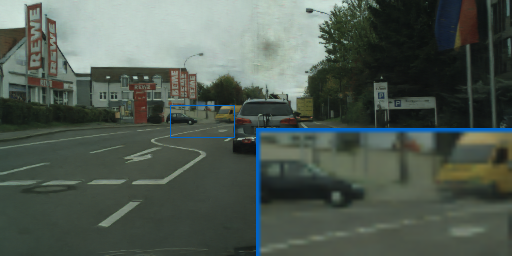}
     \includegraphics[height=0.7in]{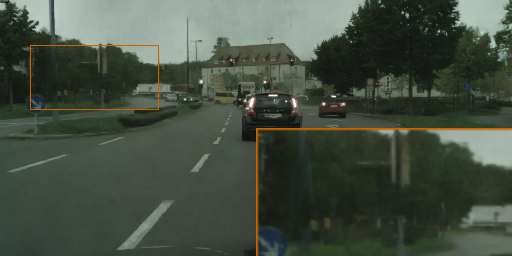}
     \includegraphics[height=0.7in]{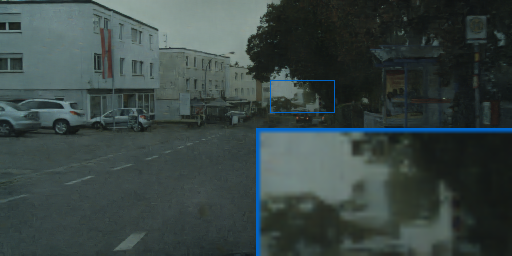}
     \includegraphics[height=0.7in]{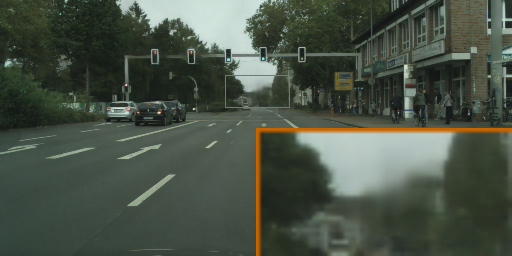}
     \includegraphics[height=0.7in]{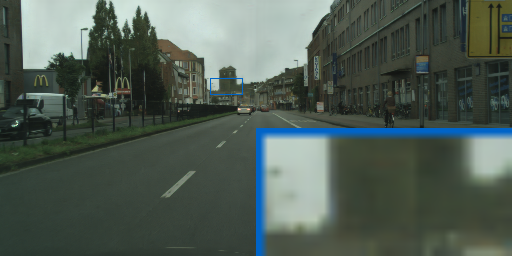}
     
     \vspace{-1.2mm}
     \renewcommand\arraystretch{0.5}
     \setlength{\tabcolsep}{7.8mm}{
     	\begin{tabular}{ccccc}
     		26.21 / 0.9485 & 28.29 / 0.9402 & 28.08 / 0.9320 & 32.68 / 0.9801 & 29.92 / 0.9863\\
     	\end{tabular}
     }\vspace{-4.8mm}

     \vspace{1mm}
     \includegraphics[height=0.7in]{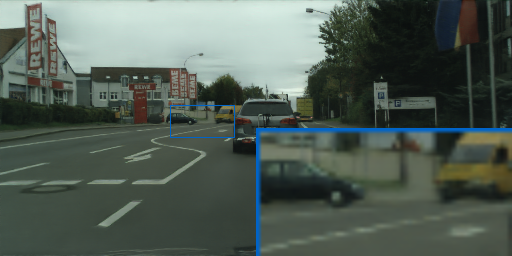}
     \includegraphics[height=0.7in]{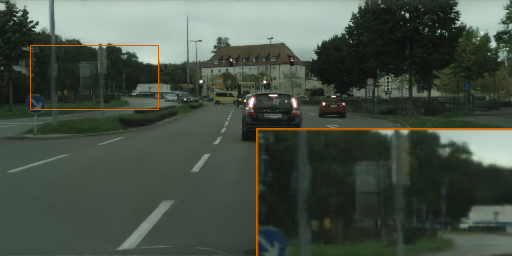}
     \includegraphics[height=0.7in]{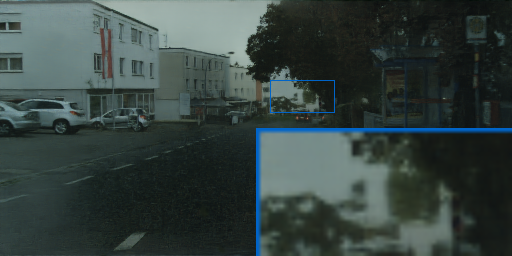}
     \includegraphics[height=0.7in]{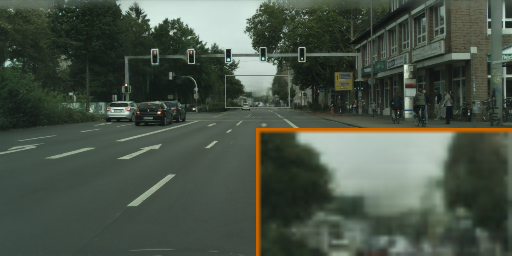}
     \includegraphics[height=0.7in]{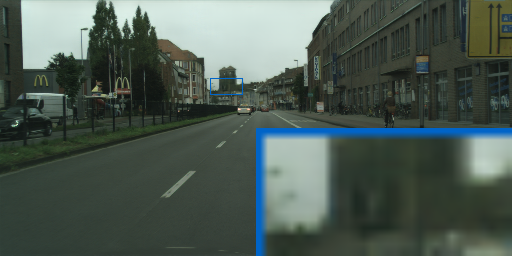}
     
     \vspace{-1.2mm}
     \renewcommand\arraystretch{0.5}
     \setlength{\tabcolsep}{7.8mm}{
     	\begin{tabular}{ccccc}
     		23.83 / 0.9540 & 28.66 / 0.9543 & 24.35 / 0.8952 & 30.70 / 0.9788 & 34.48 / 0.9886\\
     	\end{tabular}
     }\vspace{-4.8mm}

     \vspace{1mm}
     \includegraphics[height=0.7in]{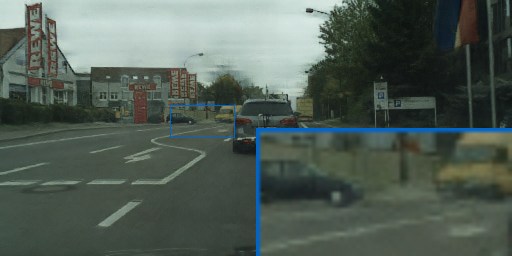}
     \includegraphics[height=0.7in]{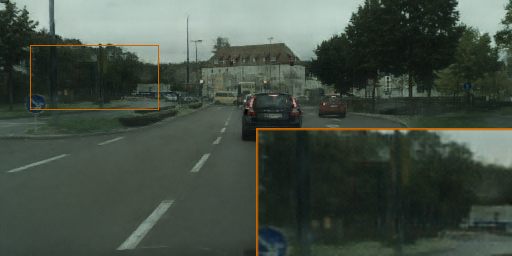}
     \includegraphics[height=0.7in]{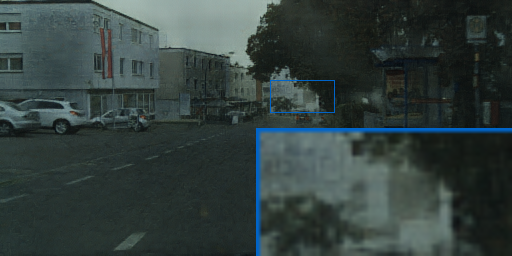}
     \includegraphics[height=0.7in]{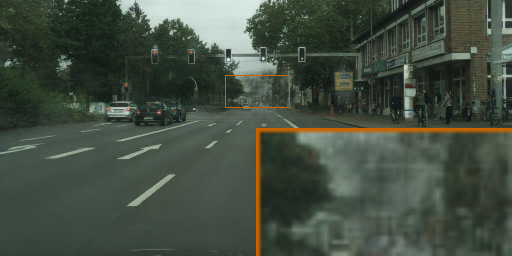}
     \includegraphics[height=0.7in]{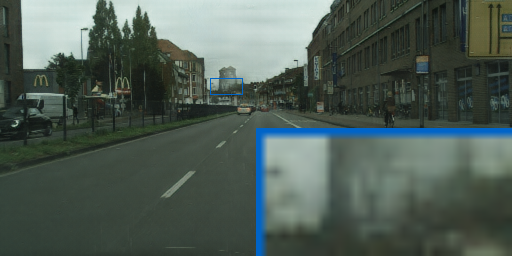}
     
     \vspace{-1.2mm}
     \renewcommand\arraystretch{0.5}
     \setlength{\tabcolsep}{7.8mm}{
     	\begin{tabular}{ccccc}
     		23.83 / 0.9169 & 26.41 / 0.9095 & 25.54 / 0.9012 & 26.99 / 0.9462 & 29.72 / 0.9670\\
     	\end{tabular}
     }\vspace{-4.8mm}

     \vspace{1mm}
     \includegraphics[height=0.7in]{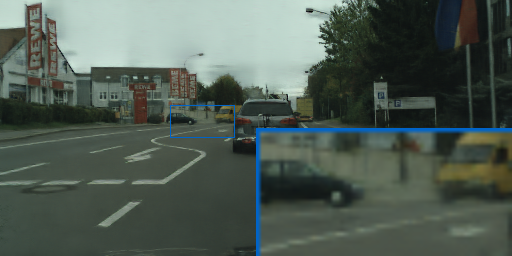}
     \includegraphics[height=0.7in]{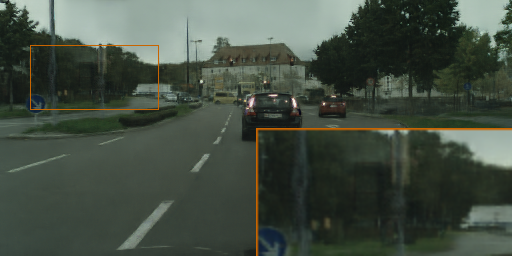}
     \includegraphics[height=0.7in]{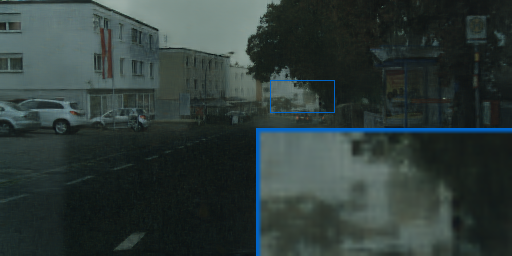}
     \includegraphics[height=0.7in]{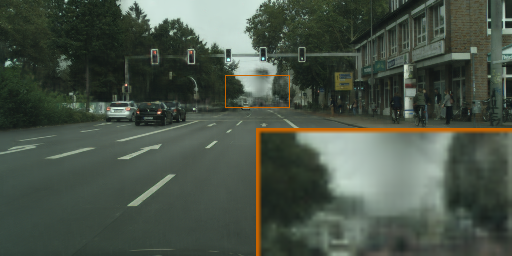}
     \includegraphics[height=0.7in]{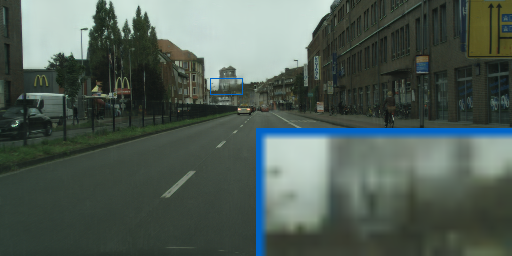}
     
     \vspace{-1.2mm}
     \renewcommand\arraystretch{0.5}
     \setlength{\tabcolsep}{7.8mm}{
     	\begin{tabular}{ccccc}
     		24.80 / 0.9444 & 27.64 / 0.9364 & 22.56 / 0.8592 & 29.03 / 0.9710 & 31.30 / 0.9832\\
     	\end{tabular}
     }\vspace{-4.8mm}

     \vspace{1mm}
     \includegraphics[height=0.7in]{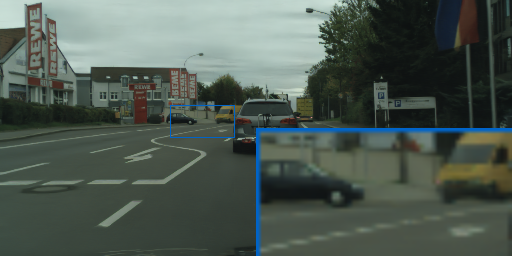}
     \includegraphics[height=0.7in]{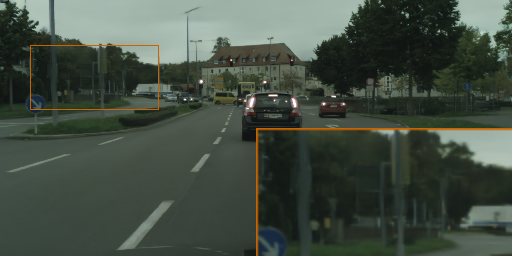}
     \includegraphics[height=0.7in]{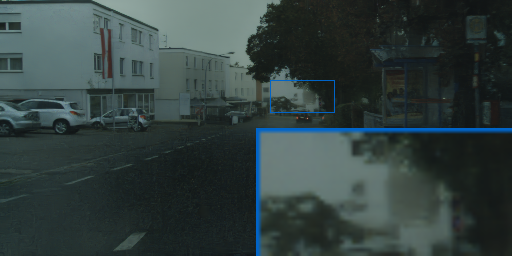}
     \includegraphics[height=0.7in]{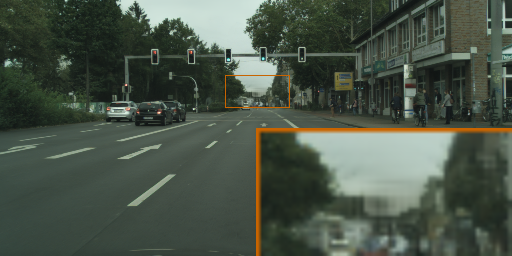}
     \includegraphics[height=0.7in]{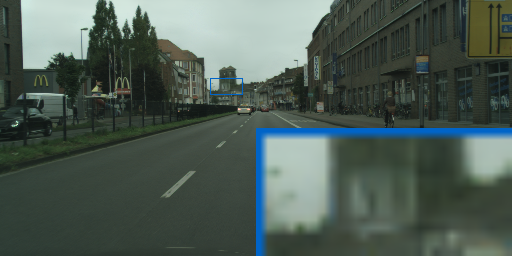}
     
     \vspace{-1.2mm}
     \renewcommand\arraystretch{0.5}
     \setlength{\tabcolsep}{7.8mm}{
     	\begin{tabular}{ccccc}
     		27.56 / 0.9621 & 30.35 / 0.9582 & 24.93 / 0.9038 & 32.92 / 0.9868 & 32.53 / 0.9913\\
     	\end{tabular}
     }\vspace{-4.8mm}

     \vspace{1mm}
     \includegraphics[height=0.7in]{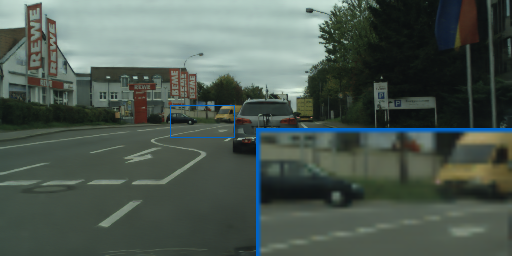}
     \includegraphics[height=0.7in]{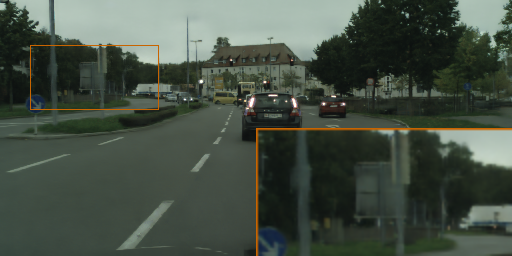}
     \includegraphics[height=0.7in]{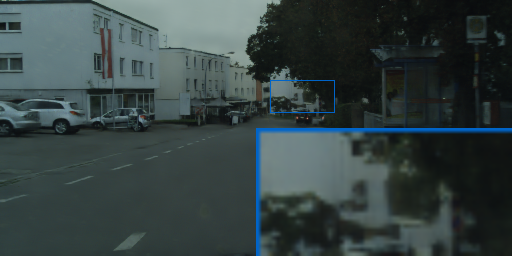}
     \includegraphics[height=0.7in]{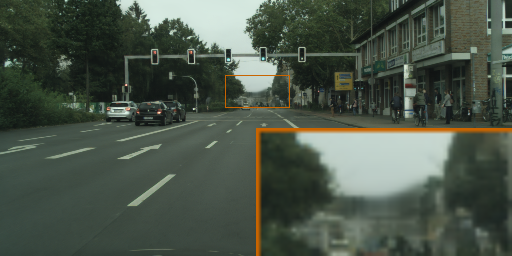}
     \includegraphics[height=0.7in]{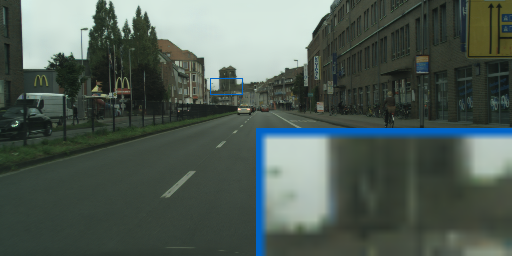}
     
     \vspace{-1.2mm}
     \renewcommand\arraystretch{0.5}
     \setlength{\tabcolsep}{7.8mm}{
     	\begin{tabular}{ccccc}
     		29.85 / 0.9710 & 31.76 / 0.9657 & 30.57 / 0.9644 & 35.83 / 0.9898 & 37.72 / 0.9927\\
     	\end{tabular}
     }\vspace{-4.8mm}

     \vspace{1mm}
     \includegraphics[height=0.7in]{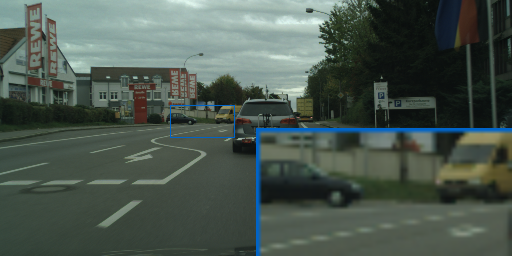}
     \includegraphics[height=0.7in]{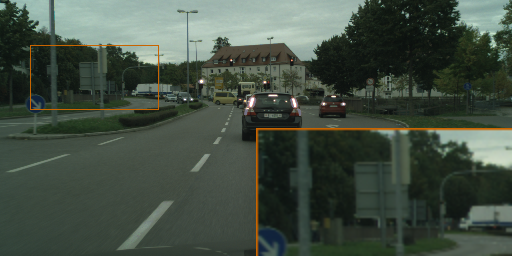}
     \includegraphics[height=0.7in]{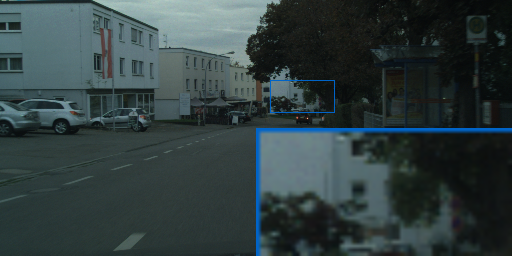}
     \includegraphics[height=0.7in]{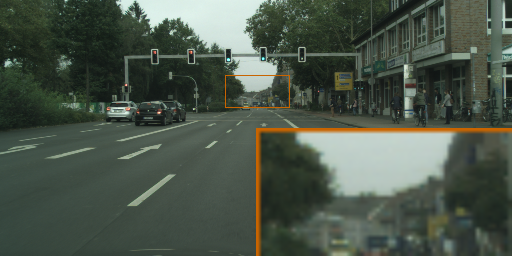}
     \includegraphics[height=0.7in]{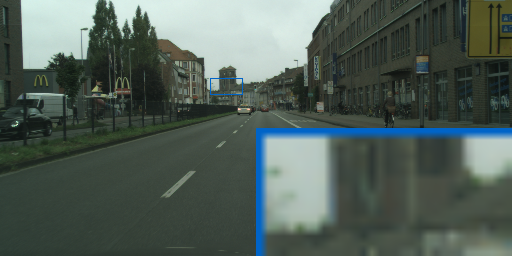}

     \caption{\textbf{Visual comparisons on RScityScape dataset}. From top to bottom: Input, DesnowNet \cite{liu2018desnownet}, PReNet\cite{ren2019progressive}, MPRNet\cite{zamir2021multi}, Uformer\cite{wang2022uformer}, TransWeather\cite{valanarasu2022transweather}, SnowFormer\cite{chen2022snowformer}, Restormer\cite{zamir2022restormer}, \textbf{RSFormer (ours)} and Ground Truth.}
     \label{fig:rscityscape}
 \end{figure*}

 \begin{figure*}[!t]
	\includegraphics[height=0.74in]{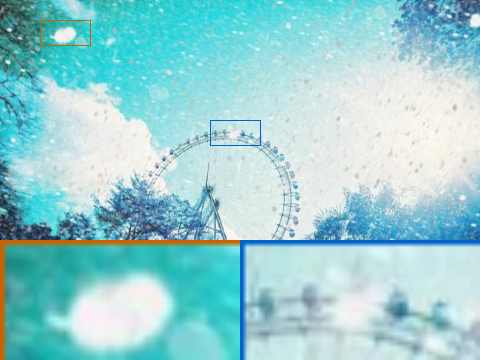}
	\includegraphics[height=0.74in]{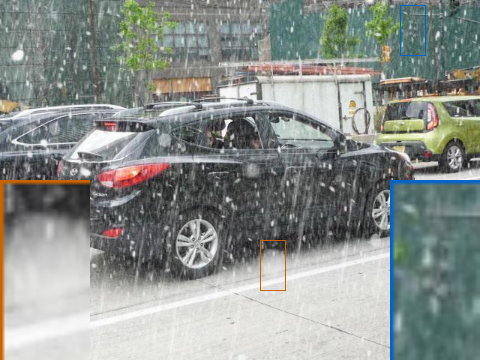}
	\includegraphics[height=0.74in]{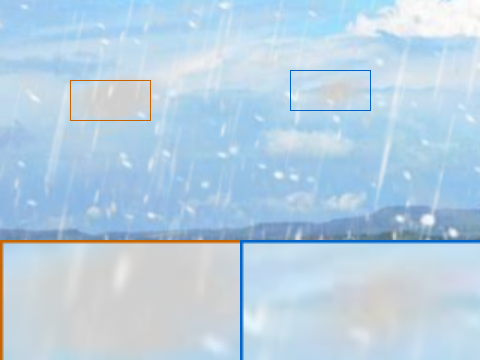}
	\includegraphics[height=0.74in]{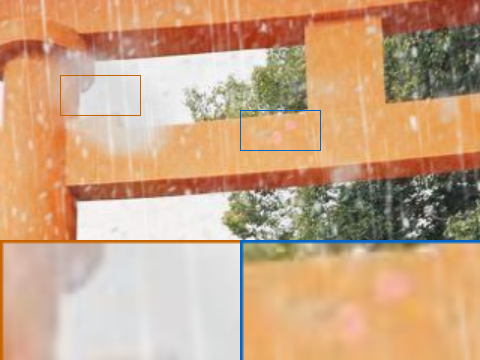}
	\includegraphics[height=0.74in]{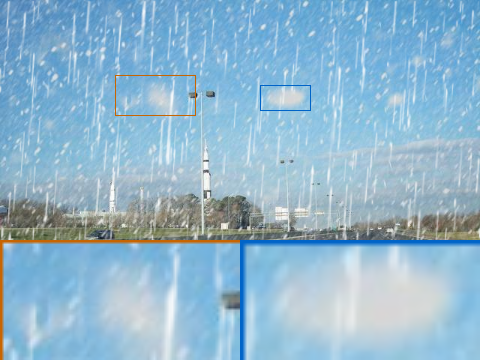}
	\includegraphics[height=0.74in]{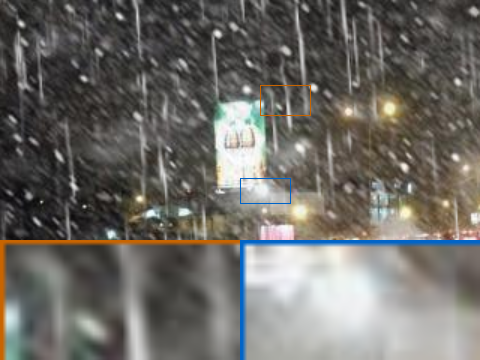}
	\includegraphics[height=0.74in]{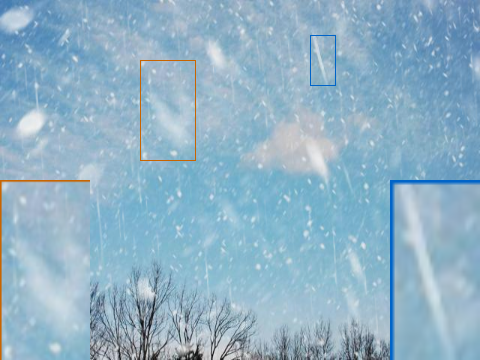}
	
        \vspace{-0.5mm}
	\renewcommand\arraystretch{0.5}
	\setlength{\tabcolsep}{2.45mm}{
		\begin{tabular}{ccccccc}
			18.11 / 0.7344 & 17.75 / 0.7629 & 21.87 / 0.8326 & 17.50 / 0.7315 & 16.54 / 0.6561 & 13.89 / 0.4571 & 24.31 / 0.8936\\
		\end{tabular}
	}\vspace{-4mm}
	
	\vspace{1mm}
	\includegraphics[height=0.74in]{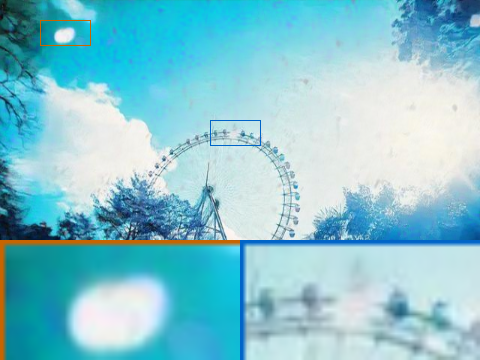}
	\includegraphics[height=0.74in]{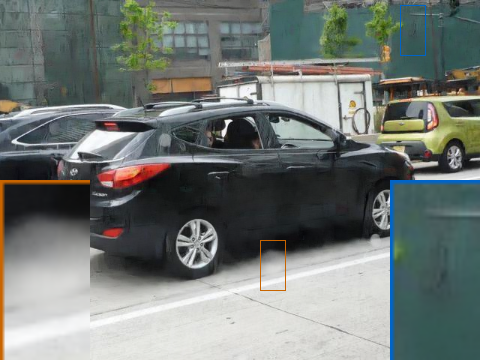}
	\includegraphics[height=0.74in]{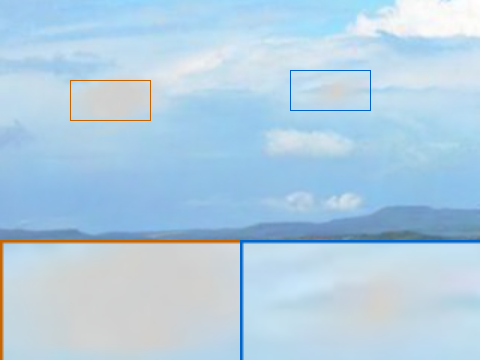}
	\includegraphics[height=0.74in]{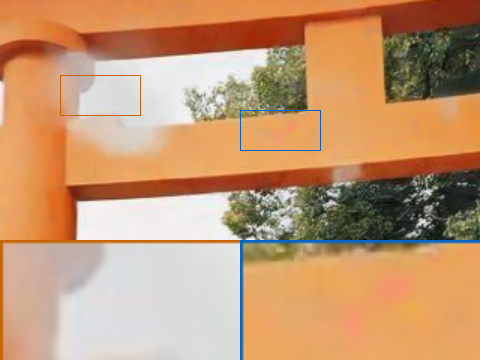}
	\includegraphics[height=0.74in]{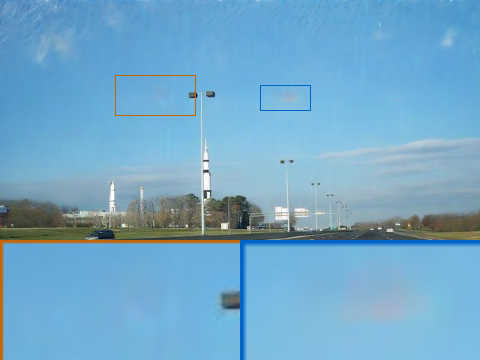}
	\includegraphics[height=0.74in]{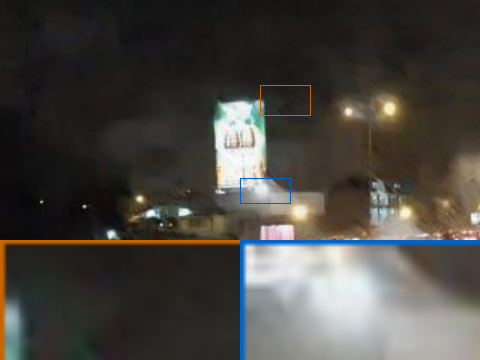}
	\includegraphics[height=0.74in]{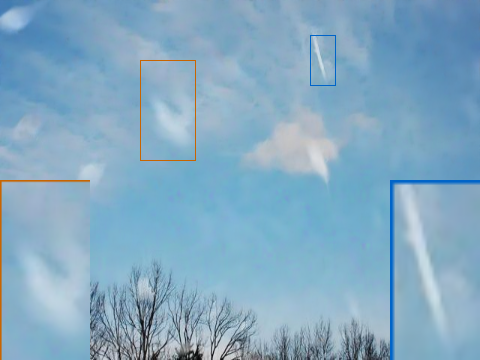}
	
        \vspace{-0.5mm}
	\renewcommand\arraystretch{0.5}
	\setlength{\tabcolsep}{2.45mm}{
		\begin{tabular}{ccccccc}
			25.34 / 0.8977 & 27.93 / 0.7629 & 31.22 / 0.9661 & 24.56 / 0.9152 & 25.04 / 0.9327 & 24.75 / 0.8874 & 30.20 / 0.9720\\
		\end{tabular}
	}\vspace{-4mm}
	
	\vspace{1mm}
	\includegraphics[height=0.74in]{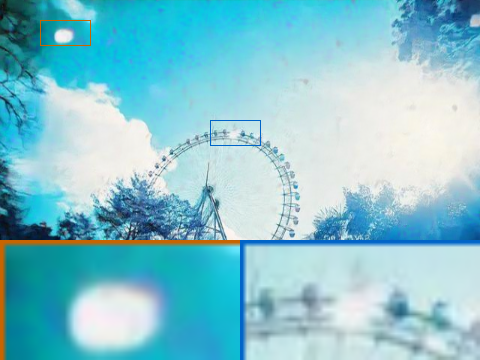}
	\includegraphics[height=0.74in]{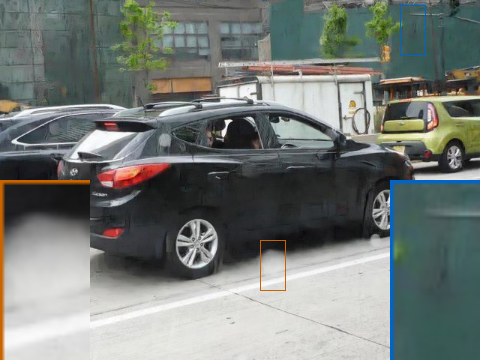}
	\includegraphics[height=0.74in]{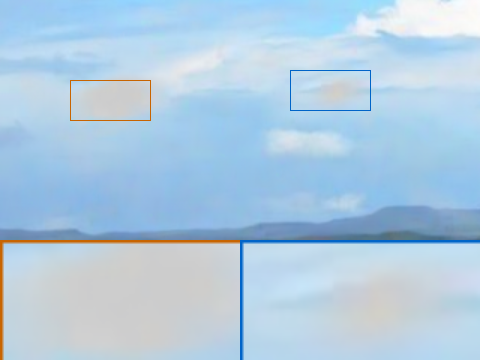}
	\includegraphics[height=0.74in]{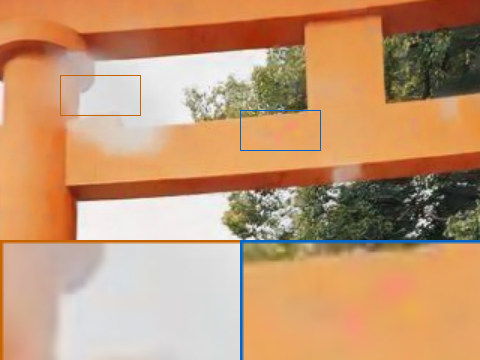}
	\includegraphics[height=0.74in]{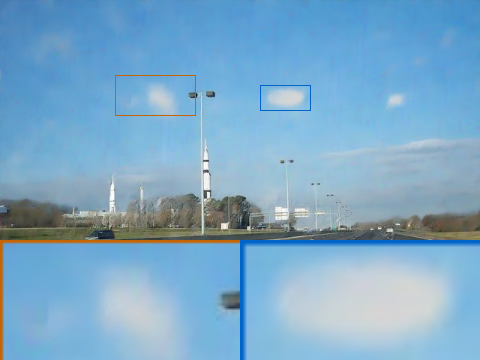}
	\includegraphics[height=0.74in]{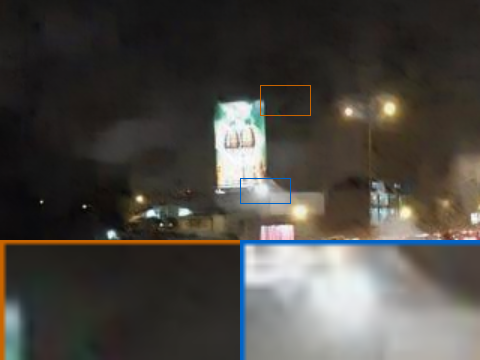}
	\includegraphics[height=0.74in]{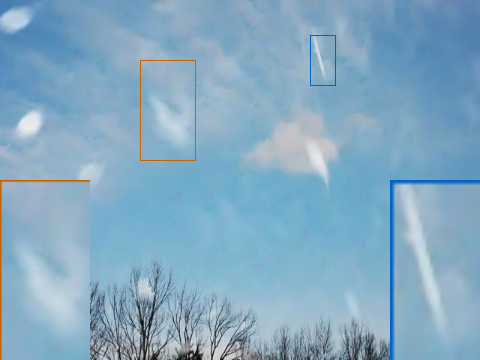}
	
        \vspace{-0.5mm}
	\renewcommand\arraystretch{0.5}
	\setlength{\tabcolsep}{2.45mm}{
		\begin{tabular}{ccccccc}
			25.39 / 0.8947 & 27.31 / 0.9250 & 31.34 / 0.9502 & 24.29 / 0.9084 & 27.68 / 0.9576 & 24.28 / 0.8774 & 28.94 / 0.9690\\
		\end{tabular}
	}\vspace{-4mm}
	
	\vspace{1mm}
	\includegraphics[height=0.74in]{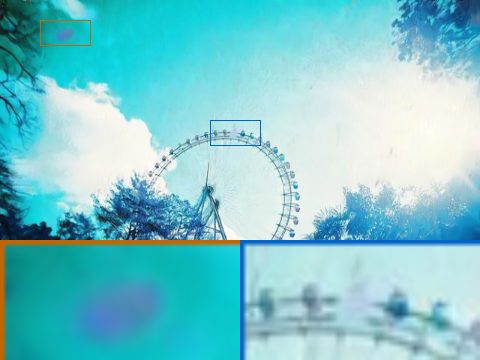}
	\includegraphics[height=0.74in]{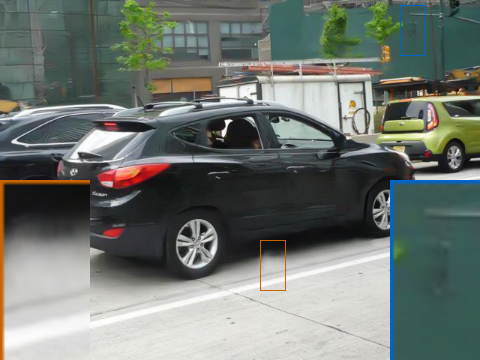}
	\includegraphics[height=0.74in]{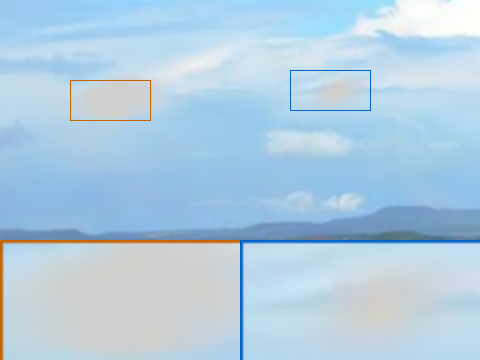}
	\includegraphics[height=0.74in]{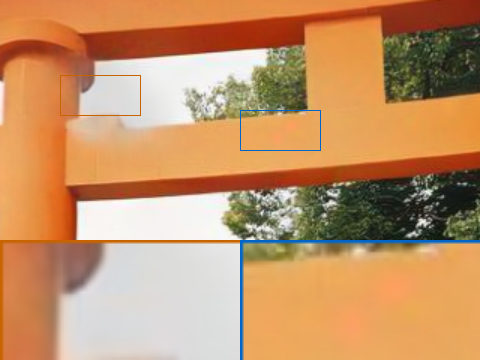}
	\includegraphics[height=0.74in]{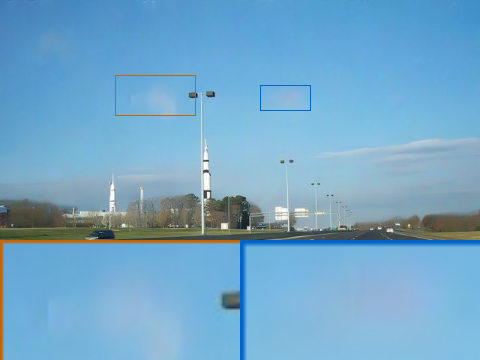}
	\includegraphics[height=0.74in]{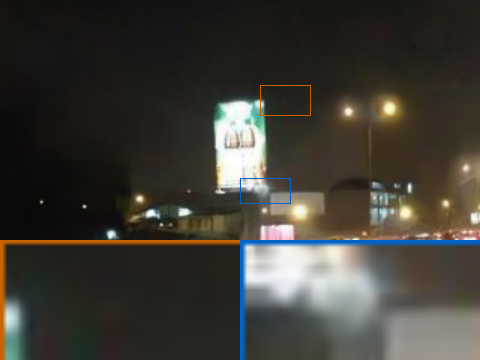}
	\includegraphics[height=0.74in]{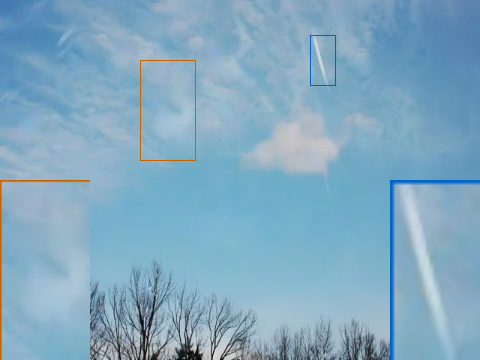}

        \vspace{-0.5mm}
	\renewcommand\arraystretch{0.5}
	\setlength{\tabcolsep}{2.45mm}{
		\begin{tabular}{ccccccc}
			28.43 / 0.9249 & 29.66 / 0.9453 & 34.11 / 0.9684 & 28.17 / 0.9405 & 33.34 / 0.9741 & 28.80 / 0.9367 & 33.54 / 0.9800\\
		\end{tabular}
	}\vspace{-4mm}
	
	\vspace{1mm}
	\includegraphics[height=0.74in]{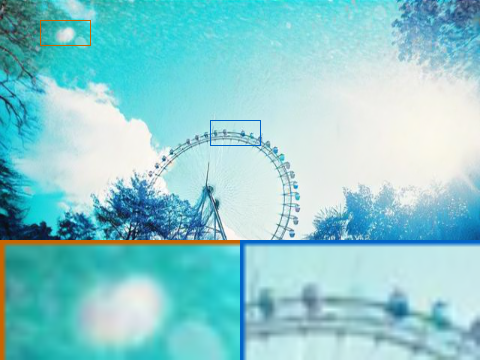}
	\includegraphics[height=0.74in]{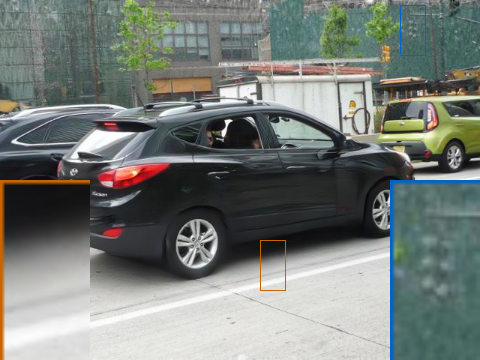}
	\includegraphics[height=0.74in]{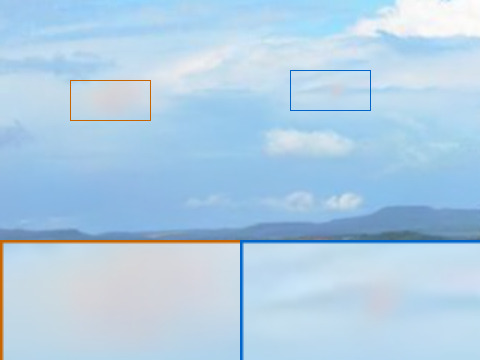}
	\includegraphics[height=0.74in]{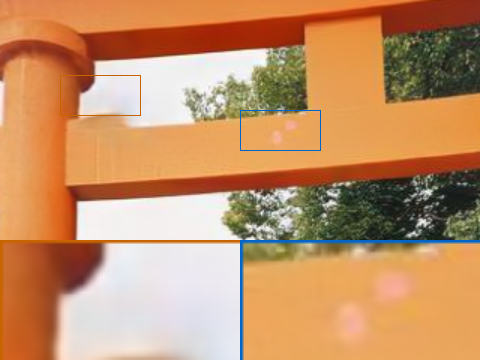}
	\includegraphics[height=0.74in]{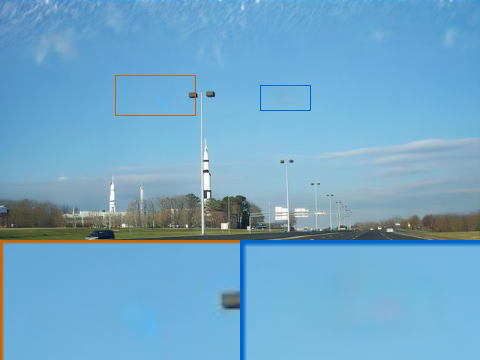}
	\includegraphics[height=0.74in]{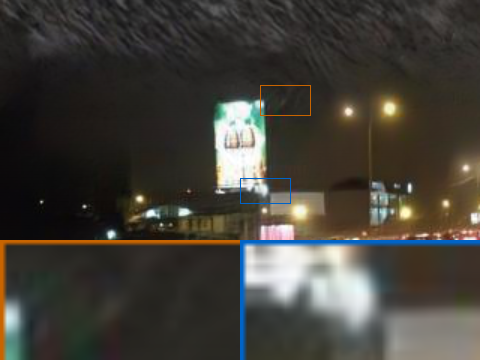}
	\includegraphics[height=0.74in]{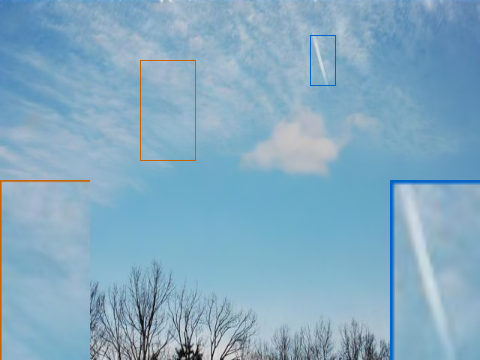}

        \vspace{-0.5mm}
	\renewcommand\arraystretch{0.5}
	\setlength{\tabcolsep}{2.45mm}{
		\begin{tabular}{ccccccc}
			22.28 / 0.8336 & 27.15 / 0.9395 & 32.17 / 0.9698 & 26.00 / 0.9259 & 25.14 / 0.9321 & 21.70 / 0.8150 & 32.43 / 0.9813\\
		\end{tabular}
	}\vspace{-4mm}
	
	\vspace{1mm}
	\includegraphics[height=0.74in]{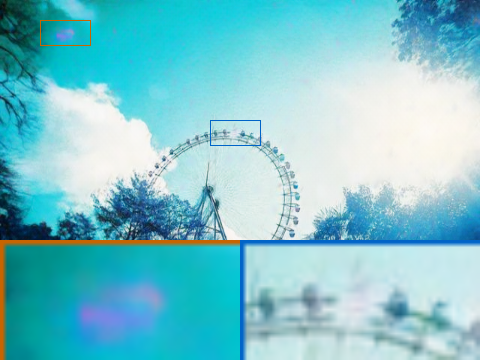}
	\includegraphics[height=0.74in]{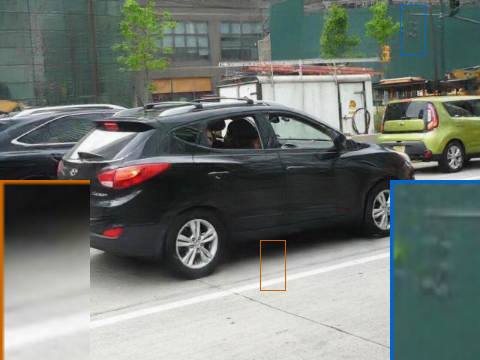}
	\includegraphics[height=0.74in]{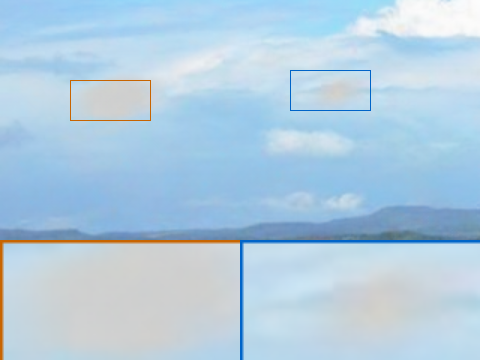}
	\includegraphics[height=0.74in]{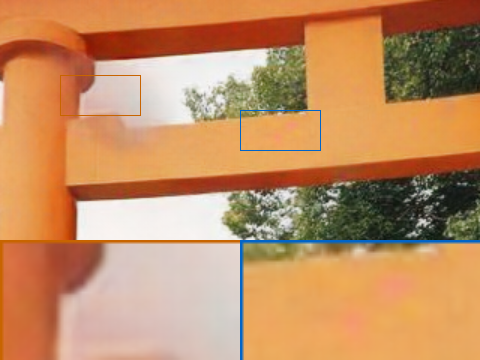}
	\includegraphics[height=0.74in]{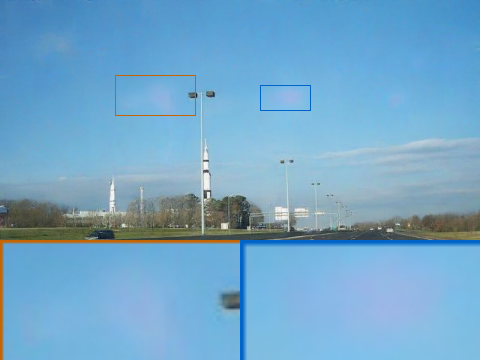}
	\includegraphics[height=0.74in]{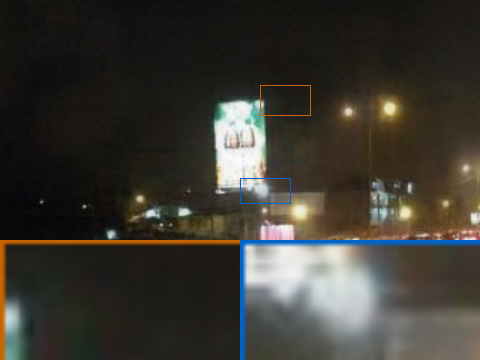}
	\includegraphics[height=0.74in]{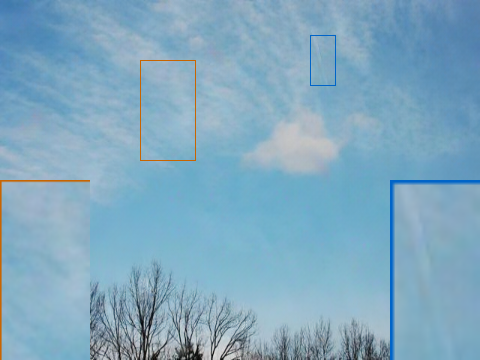}

        \vspace{-0.5mm}
	\renewcommand\arraystretch{0.5}
	\setlength{\tabcolsep}{2.45mm}{
		\begin{tabular}{ccccccc}
			27.65 / 0.9107 & 29.26 / 0.9378 & 32.37 / 0.9628 & 27.59 / 0.9315 & 32.91 / 0.9678 & 28.34 / 0.9236 & 33.73 / 0.9793\\
		\end{tabular}
	}\vspace{-4mm}
	
	\vspace{1mm}
	\includegraphics[height=0.74in]{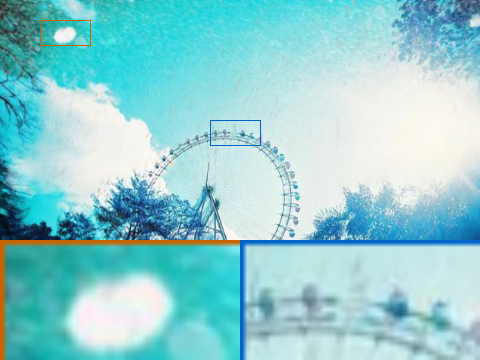}
	\includegraphics[height=0.74in]{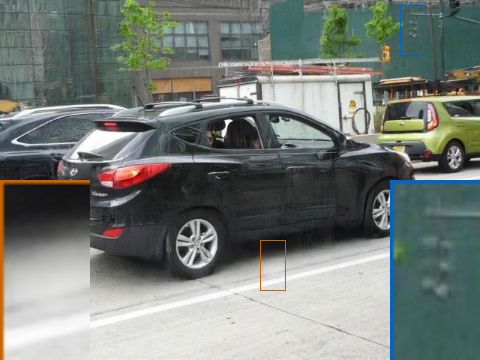}
	\includegraphics[height=0.74in]{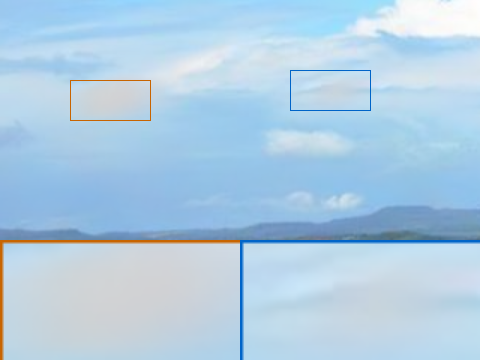}
	\includegraphics[height=0.74in]{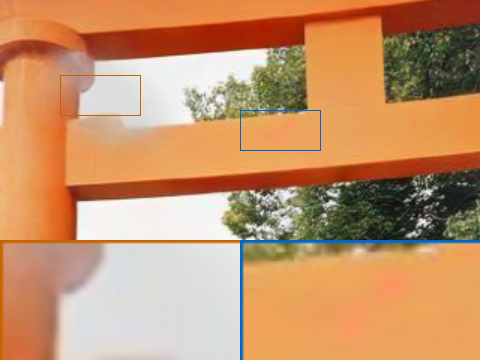}
	\includegraphics[height=0.74in]{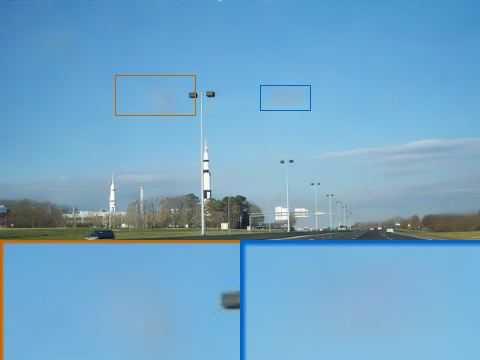}
	\includegraphics[height=0.74in]{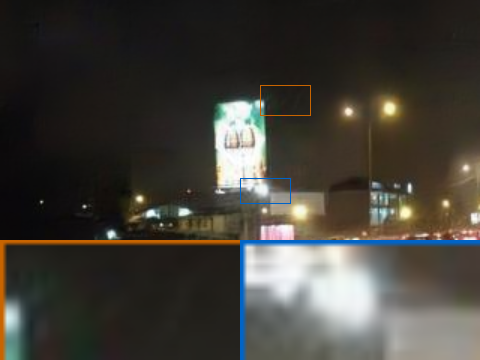}
	\includegraphics[height=0.74in]{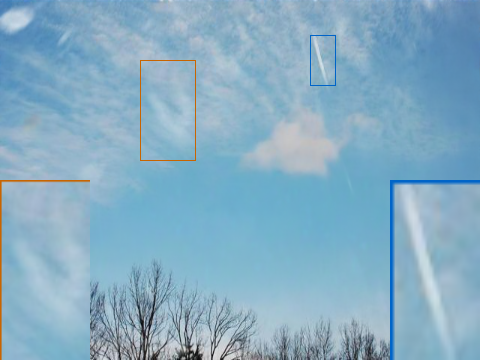}

        \vspace{-0.5mm}
	\renewcommand\arraystretch{0.5}
	\setlength{\tabcolsep}{2.45mm}{
		\begin{tabular}{ccccccc}
			21.56 / 0.8243 & 28.50 / 0.9392 & 34.42 / 0.9743 & 24.48 / 0.9135 & 33.97 / 0.9786 & 29.43 / 0.9450 & 29.73 / 0.9768\\
		\end{tabular}
	}\vspace{-4mm}
	
	\vspace{1mm}
	\includegraphics[height=0.74in]{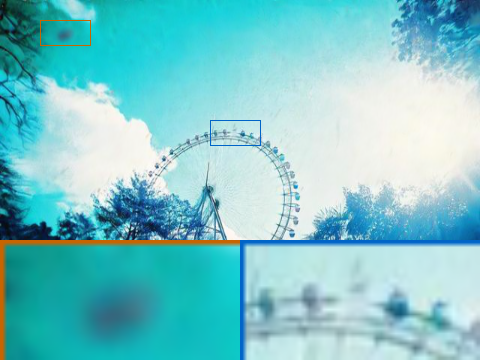}
	\includegraphics[height=0.74in]{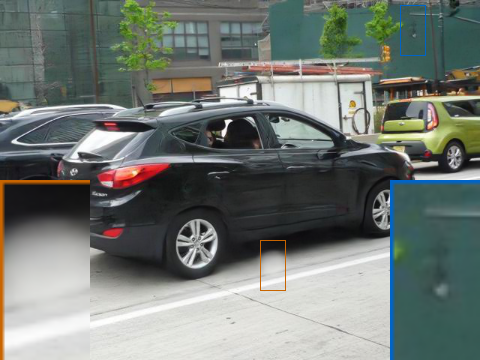}
	\includegraphics[height=0.74in]{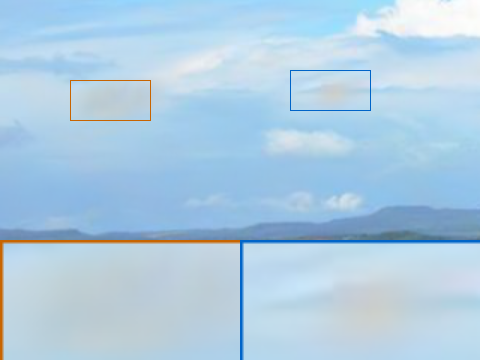}
	\includegraphics[height=0.74in]{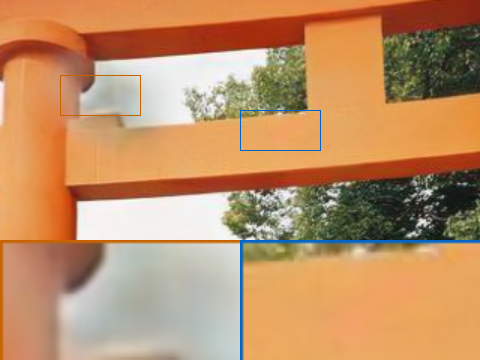}
	\includegraphics[height=0.74in]{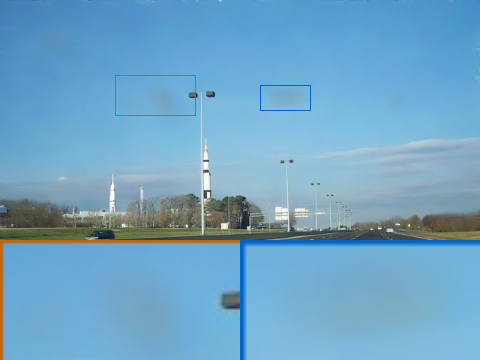}
	\includegraphics[height=0.74in]{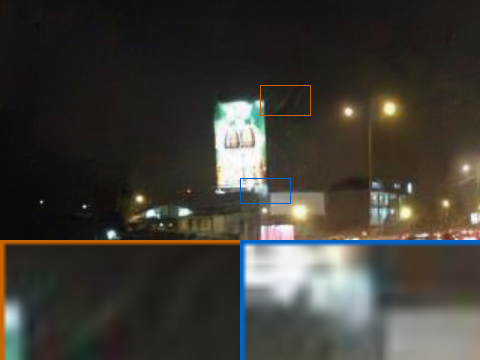}
	\includegraphics[height=0.74in]{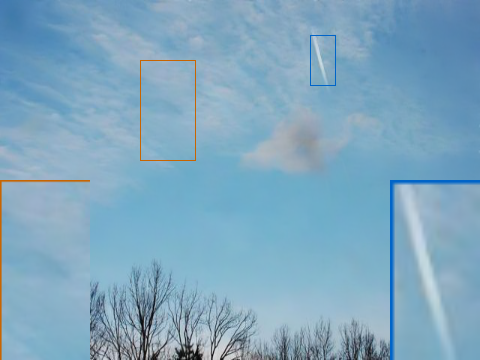}

        \vspace{-0.5mm}
	\renewcommand\arraystretch{0.5}
	\setlength{\tabcolsep}{2.45mm}{
		\begin{tabular}{ccccccc}
			28.92 / 0.9299 & 30.22 / 0.9512 & 33.08 / 0.9714 & 28.11 / 0.9476 & 33.87 / 0.9778 & 28.55 / 0.9383 & 32.87 / 0.9825\\
		\end{tabular}
	}\vspace{-4mm}
	
	\vspace{1mm}
	\includegraphics[height=0.74in]{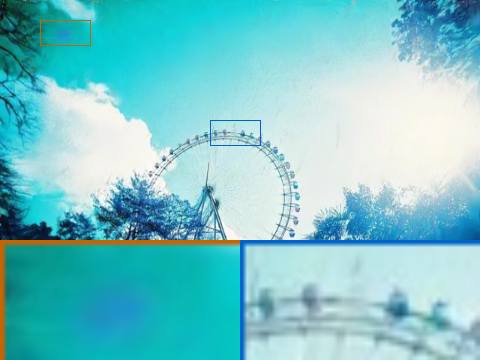}
	\includegraphics[height=0.74in]{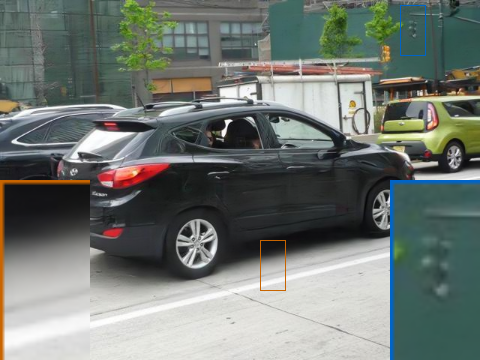}
	\includegraphics[height=0.74in]{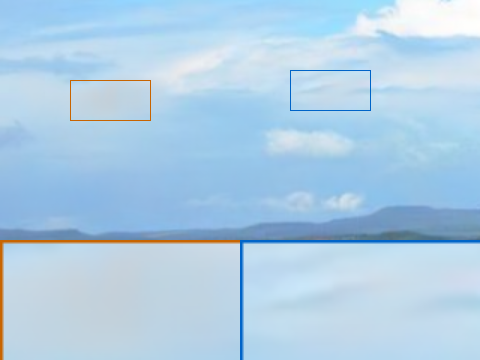}
	\includegraphics[height=0.74in]{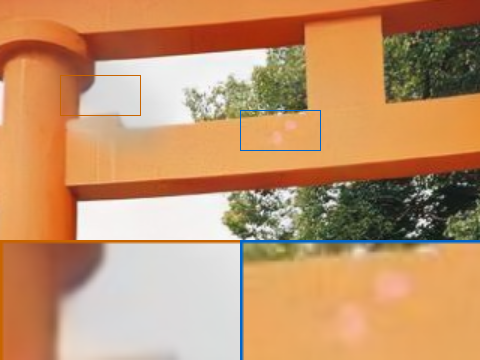}
	\includegraphics[height=0.74in]{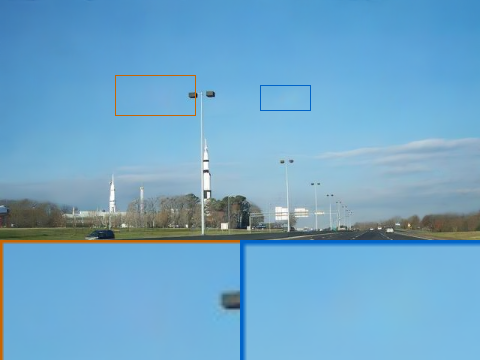}
	\includegraphics[height=0.74in]{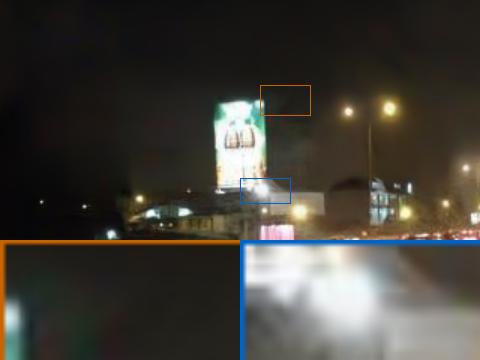}
	\includegraphics[height=0.74in]{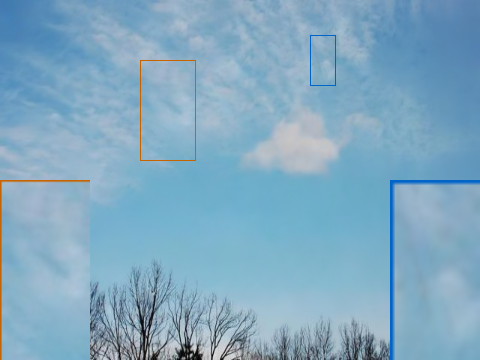}

        \vspace{-0.5mm}
	\renewcommand\arraystretch{0.5}
	\setlength{\tabcolsep}{2.45mm}{
		\begin{tabular}{ccccccc}
			29.31 / 0.9373 & 30.87 / 0.9540 & 34.83 / 0.9714 & 28.67 / 0.9492 & 33.96 / 0.9768 & 28.98 / 0.9403 & 34.57 / 0.9835\\
		\end{tabular}
	}\vspace{-4mm}
	
	\vspace{1mm}
	\includegraphics[height=0.74in]{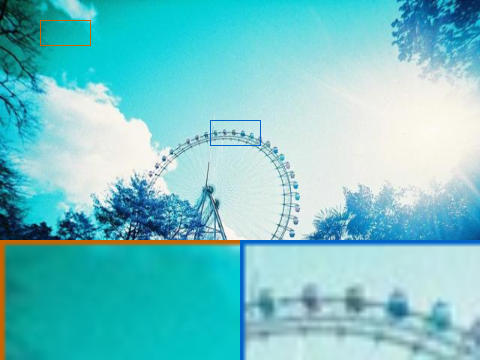}
	\includegraphics[height=0.74in]{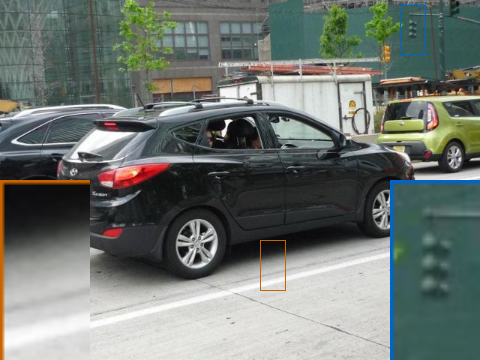}
	\includegraphics[height=0.74in]{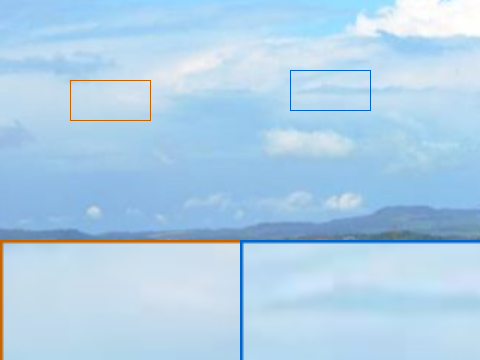}
	\includegraphics[height=0.74in]{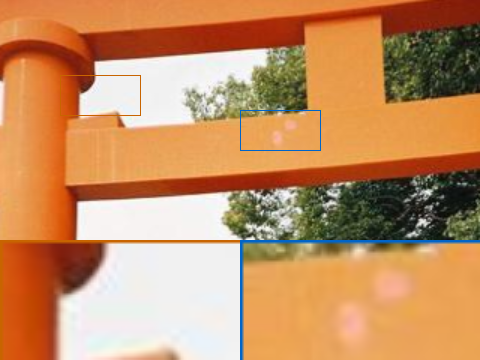}
	\includegraphics[height=0.74in]{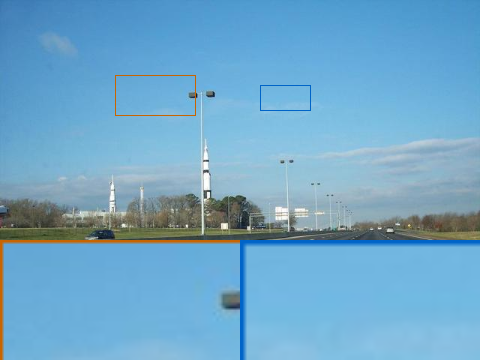}
	\includegraphics[height=0.74in]{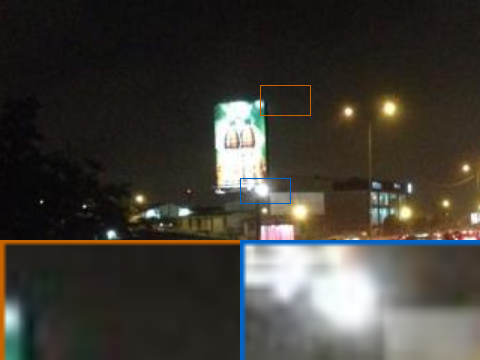}
	\includegraphics[height=0.74in]{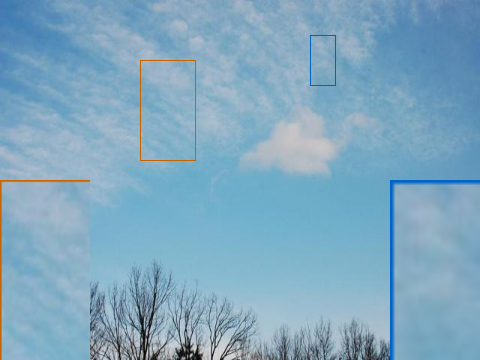}
	
	\caption{\textbf{Visual comparisons on RS100K dataset}. From top to bottom: Input, DesnowNet \cite{liu2018desnownet}, PReNet\cite{ren2019progressive}, MPRNet\cite{zamir2021multi}, Uformer\cite{wang2022uformer}, TransWeather\cite{valanarasu2022transweather}, SnowFormer\cite{chen2022snowformer}, Restormer\cite{zamir2022restormer}, \textbf{RSFormer (ours)} and Ground Truth.}
	\label{fig:rs100k}
\end{figure*}

\subsection{Single Weather Removal}
To further evaluate our proposed RSFormer in single weather removal, we conduct experiments on RainCityScape \cite{hu2019depth} dataset and Snow100K \cite{liu2018desnownet} dataset.
TABLE \ref{tab:RainCityScape} shows the quantitative comparisons on RainCityScape dataset, where NANAS achieves the best performance in PSNR while our proposed RSFormer obtains the highest score in SSIM.
However, our RSFormer obtains a performance decrease of 0.03 dB comparing with NANAS \cite{cai2022multi} in PSNR.
Therefore, RSFormer is not much competitive on RainCityScape dataset over MANAS.
TABLE \ref{tab:Snow100K} shows that our RSFormer achieves the best performance on Snow100K dataset.
Specifically, RSFormer obtains performance gains of 0.99 dB, 0.73 dB and 0.92 dB over SnowFormer on Snow100K-L, Snow100K-M and Snow100K-S dataset respectively. Therefore, RSFormer is not much competitive on RainCityScape dataset, however it achieves the state-of-the-art performance on Snow100K dataset.

\begin{table*}[!t]
\begin{minipage}{0.37\linewidth}
     \centering
     \caption{Image deraining results. Our RSFormer achieves the best performance in SSIM on RainCityScape \cite{hu2019depth} dataset.}
     \label{tab:RainCityScape}
     \renewcommand\arraystretch{1.25}
     \setlength{\tabcolsep}{1.5mm}{
     \begin{tabular}{lccc}
     \toprule
     Method & Venue & PSNR & SSIM \\
    \midrule
    DIDMDN \cite{zhang2018density} & \textit{CVPR} 2018 & 28.43 & 0.9349 \\
    RESCAN \cite{li2018recurrent} & \textit{ECCV} 2018 & 24.49 & 0.8852 \\
    DCPDN \cite{zhang2018densely} & \textit{CVPR} 2018 & 28.52 & 0.9277\\
    DAFNet \cite{hu2019depth} & \textit{CVPR} 2019 & 30.06 & 0.9530 \\
    DRDNet \cite{deng2020detail} & \textit{CVPR} 2020 & 30.13 & 0.9535 \\
    MPRNet \cite{zamir2021multi} & \textit{CVPR} 2021 & 30.96 & 0.9721 \\
    EPRRNet \cite{zhang2022beyond} & \textit{IJCV} 2022 & 31.11 & 0.9741 \\
    Uformer \cite{wang2022uformer} & \textit{CVPR} 2022 & 33.38 & 0.9839 \\
    Restormer \cite{zamir2022restormer} & \textit{CVPR} 2022 & 35.08 & \underline{0.9905}\\
    MANAS \cite{cai2022multi} & \textit{TCSVT} 2022 & \textbf{35.19} & 0.9840 \\
    \textbf{RSFormer} & --- & \underline{35.16} & \textbf{0.9919}\\
    \bottomrule
    \end{tabular}
     }
\end{minipage}
\hfill
\begin{minipage}{0.6\linewidth}
     \centering
     \caption{Image desnowing results. Our RSFormer achieves the best performance on Snow100K \cite{liu2018desnownet} dataset.}
     \label{tab:Snow100K}
     \renewcommand\arraystretch{1.25}
     \setlength{\tabcolsep}{1.5mm}{
     \begin{tabular}{lccccccc}
     \toprule
     \multirow{2}{*}{Method} & \multirow{2}{*}{Venue} & \multicolumn{2}{c}{Snow100K-L} & \multicolumn{2}{c}{Snow100K-M} & \multicolumn{2}{c}{Snow100K-S} \\
     ~ & ~ & PSNR & SSIM & PSNR & SSIM & PSNR & SSIM \\     
    \midrule
    DeepLab \cite{chen2017deeplab} & \textit{TPAMI} 2018 & 21.29 & 0.7747 & 24.37 & 0.8572 & 25.95 & 0.8783 \\ 
    RESCAN \cite{li2018recurrent} & \textit{CVPR} 2018 & 26.08 & 0.8108 & 29.95 & 0.8860 & 31.51 & 0.9032 \\
    DesnowNet \cite{liu2018desnownet} & \textit{TIP} 2018 & 27.17 & 0.8983 & 30.87 & 0.9490 & 32.33 & 0.9500\\
    SPANet \cite{wang2019spatial} & \textit{CVPR} 2019 & 23.70 & 0.7930 & 28.06 & 0.8680 & 29.92 & 0.8260 \\
    JSTASR \cite{chen2020jstasr} & \textit{ECCV} 2020 & 25.32 & 0.8076 & 29.11 & 0.8843 & 31.40 & 0.9012 \\
    DesnowGAN \cite{jaw2020desnowgan} & \textit{TCSVT} 2021 & 28.07 & 0.9211 & 31.88 & 0.9570 & 33.43 & 0.9641 \\
    DDMSNet \cite{zhang2021deep} & \textit{TIP} 2021 & 28.85 & 0.8772 & 32.89 & 0.9330 & 34.34 & 0.9445 \\
    TransWeather \cite{valanarasu2022transweather} & \textit{CVPR} 2022 & 29.17 & 0.9304 & 32.68 & 0.9603 & 33.94 & 0.9665 \\
    SnowFormer \cite{chen2022snowformer} & \textit{arXiV} 2022 & \underline{29.49} & \underline{0.9314} & \underline{33.71} & \underline{0.9646} & \underline{35.25} & \underline{0.9702}\\
    \textbf{RSFormer} & --- & \textbf{30.48} & \textbf{0.9464} & \textbf{34.44} & \textbf{0.9714} & \textbf{36.17} & \textbf{0.9764} \\
    \bottomrule
    \end{tabular}
     }
\end{minipage}
 \end{table*}

 \section{Ablation Study and Discussion}
 To demonstrate the effectiveness of our proposed RSFormer, we conduct ablation experiments on RSCityScape dataset. We keep the same configuration as the above description and illustrate the effectiveness of each component separately.

 \subsection{Improvements of Individual Components}
\label{sec:components} 
We start the baseline with u-shaped Transformer with identity mapping as the token mixer, where the (transposed) convolution is adopted for features sampling and $\mathcal{L}_{spat}$ loss function is utilized to train the network.
We compare and analyze the improvements of transposed self-attention, Transformer-like convolution block, convolution sampling mchanism (CSM), global-local self-attention sampling (GLASM) mechanism and spatial-frequency loss (SFL) with similar number of parameters.
Note that common spatial-wise self-attention cannot be performed in practical experimental environments, and therefore we only discuss the transposed (channel) self-attention \cite{zamir2022restormer, ding2022davit}.
As shown in TABLE \ref{tab:component}, our TCB outperforms TSA by 1.76 dB and 1.27\% in PSNR/SSIM, which indicates that our TCB achieves better intra-stage feature learning over transposed self-attention and confirms the necessity of using locality-wise global information instead of globality-wise global one during the multiple hierarchical stages.
Furthermore, our proposed GLASM obtains a significant gain of 1.02 dB over CSM for down-/up-sampling features, which demonstrates the effectiveness of our consideration of both global and local dependencies for cross-stage progression.
In the end, we adopt SFL to train our RSFormer and attain higher PSNR/SSIM scores.
Therefore, each component designed for rain-by-snow weather removal is effective to promote the quality of generated image.

\begin{table}[!t]
     \centering
     \caption{Ablation study of individual components. Each proposed component shows a positive effect on the overall performance.}
     \label{tab:component}
     \renewcommand\arraystretch{1.25}
     \setlength{\tabcolsep}{1.7mm}{
     \begin{tabular}{lcccccccc}
     \toprule
     \multirow{2}{*}{Model} & \multicolumn{5}{|c|}{Component} & \multirow{2}{*}{PSNR} & \multirow{2}{*}{SSIM} \\
     ~ & \multicolumn{1}{|c}{TSA} & TCB & CSM & GLASM & \multicolumn{1}{c|}{SFL} \\
    \midrule
    Baseline & \multicolumn{1}{|c}{~} & ~ & \checkmark & ~ & \multicolumn{1}{c|}{~} & 27.12 & 0.894 \\
    \romannumeral2 & \multicolumn{1}{|c}{\checkmark} & ~ & \checkmark & ~ & \multicolumn{1}{c|}{~} & 31.76 & 0.944 \\
    \romannumeral3 & \multicolumn{1}{|c}{~} & \checkmark & \checkmark & ~ & \multicolumn{1}{c|}{~} & 33.13 & 0.956 \\
    \romannumeral4 & \multicolumn{1}{|c}{~} & \checkmark & ~ & \checkmark & \multicolumn{1}{c|}{~} & 34.15 & 0.969 \\
    \textbf{RSFormer} & \multicolumn{1}{|c}{~} & \checkmark & ~ & \checkmark & \multicolumn{1}{c|}{\checkmark} & \textbf{34.54} & \textbf{0.981} \\
    \bottomrule
    \end{tabular}
     }
 \end{table}
 
\subsection{Attention Sampling}
To further evaluate our proposed attention-based sampling mechanism GLASM, we conduct comparative experiments of commonly used down-/up-sampling approaches, (transposed) convolution, pixel-(un)shuffle, our proposed GASM and GLASM. We report the PSNR/SSIM scores in TABLE \ref{tab:sampling}.
Overall, our proposed GLASM outperforms other usual sampling mechanisms by 1.33 dB and 1.06 dB. Without local enhancement, GASM can also outperform the second-best Restormer \cite{zamir2022restormer}.
Therefore, our GLASM is effective for cross-stage progression while providing a novel direction of sampling design in weather-degraded image restoration.

\begin{table}[!t]
     \centering
     \caption{Ablation study of attention sampling. Our proposed attention sampling mechanism achieves better performance than commonly used sampling operators.}
     \label{tab:sampling}
     \renewcommand\arraystretch{1.25}
     \setlength{\tabcolsep}{7mm}{
     \begin{tabular}{lcc}
     \toprule
    Sampling & PSNR & SSIM \\
    \midrule
    (Transposed) Convolution & 33.21 & 0.952 \\
    Pixel-(un)shuffle & 33.48 & 0.960 \\
    \textbf{GASM} & \underline{33.87} & \underline{0.974} \\
    \textbf{GLASM} & \textbf{34.54} & \textbf{0.981} \\
    \bottomrule
    \end{tabular}
     }
 \end{table}
 
 As we stated in Sec. \ref{sec:method}, transposed self-attention is lighter than spatial self-attention. Specifically, when we utilize SSA in the proposed GLASM, RSFormer can not be trained or tested with the same hyper-parameters and experimental environments to TSA-based sampling RSFormer.
 TABLE \ref{tab:glasm} shows that SSA-based sampling RSFormer obtains a performance gain of 0.93 dB in PSNR over TSA-based sampling RSFormer.
 However, although SSA-based sampling achieves better performance for rain-by-snow weather removal, it requires more computational amount and is not efficient.

 \begin{table}[H]
     \centering
     \caption{Abalation experiments of transposed and spatial self-attention sampling. The latter achieves the better performance in both PSNR and SSIM.}
     \label{tab:glasm}
     \renewcommand\arraystretch{1.25}
     \setlength{\tabcolsep}{8mm}{
     \begin{tabular}{ccc}
     \toprule
     Attention Sampling & PSNR & SSIM \\
    \midrule
    TSA-based & 34.54  & 0.981  \\
    SSA-based & \textbf{35.47} & \textbf{0.990} \\
    \bottomrule
    \end{tabular}
     }
 \end{table}

\begin{figure*}[!t]
	\includegraphics[height=0.74in]{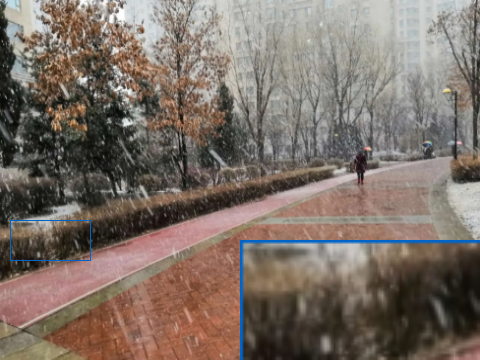}
	\includegraphics[height=0.74in]{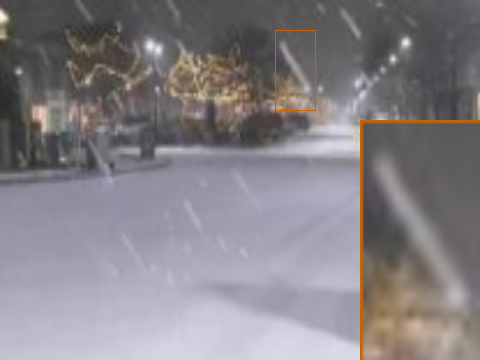}
	\includegraphics[height=0.74in]{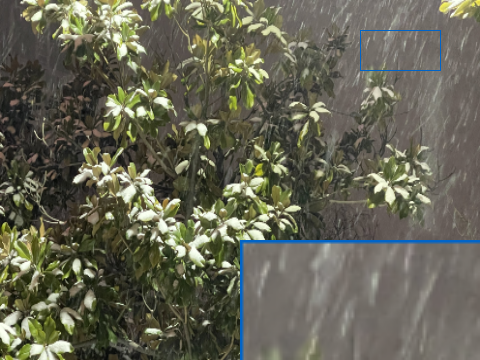}
	\includegraphics[height=0.74in]{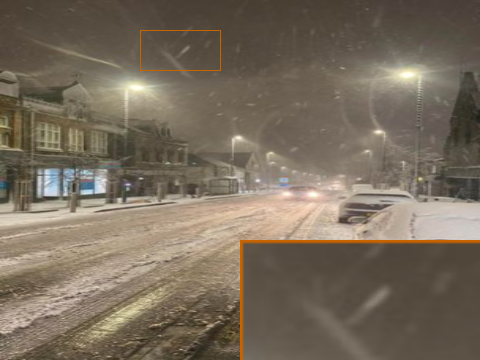}
	\includegraphics[height=0.74in]{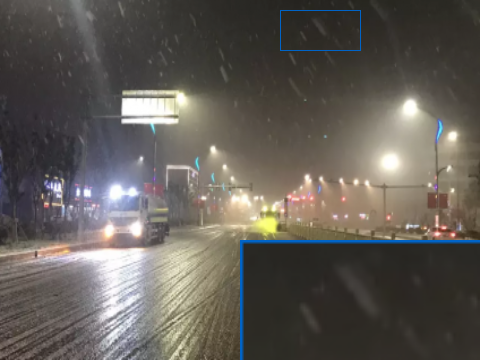}
	\includegraphics[height=0.74in]{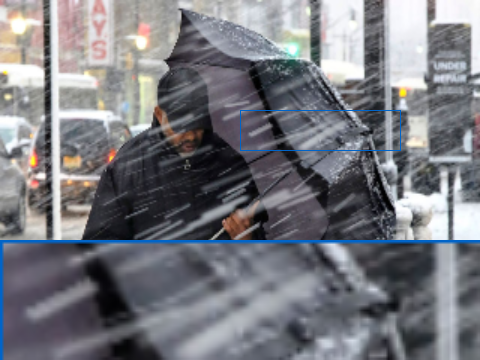}
	\includegraphics[height=0.74in]{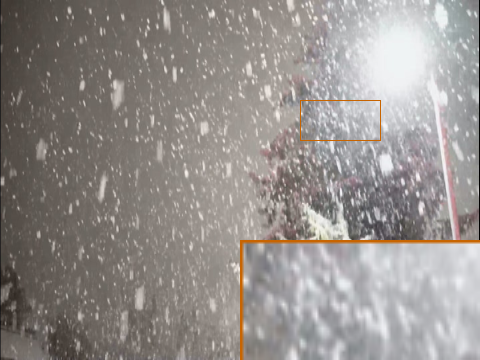}
	
	\vspace{1mm}
	\includegraphics[height=0.74in]{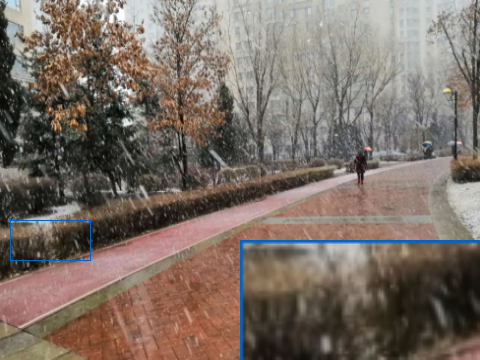}
	\includegraphics[height=0.74in]{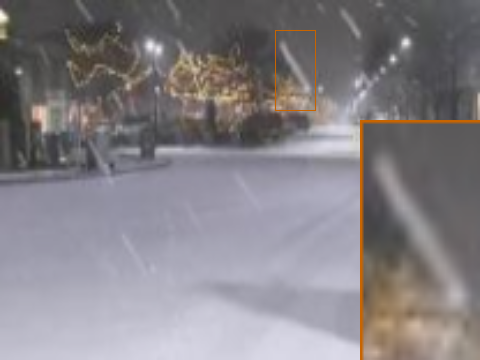}
	\includegraphics[height=0.74in]{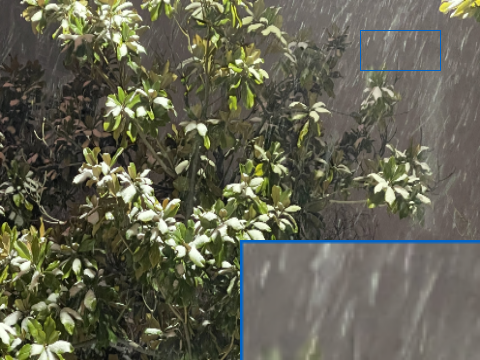}
	\includegraphics[height=0.74in]{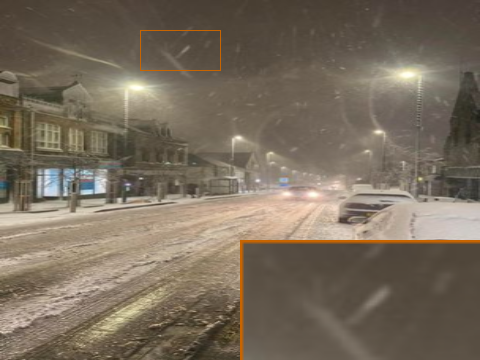}
	\includegraphics[height=0.74in]{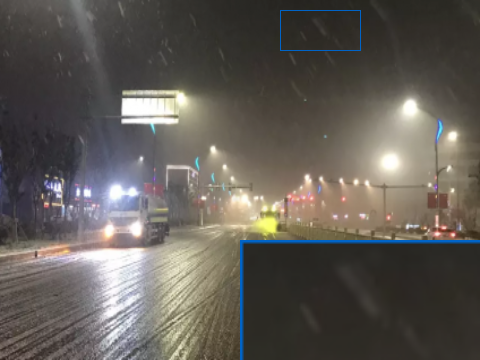}
	\includegraphics[height=0.74in]{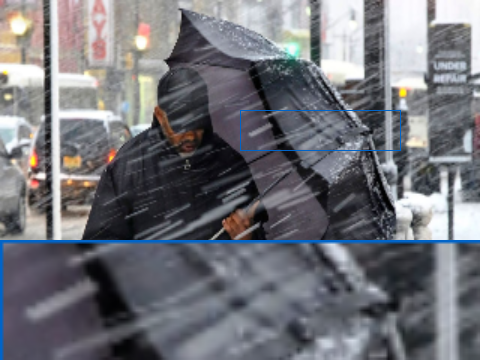}
	\includegraphics[height=0.74in]{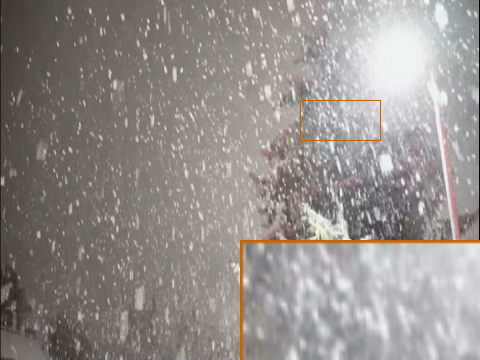}
	
	\vspace{1mm}
	\includegraphics[height=0.74in]{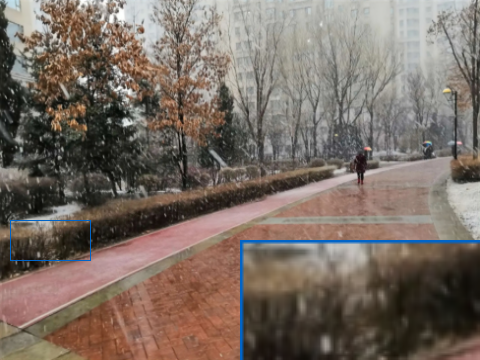}
	\includegraphics[height=0.74in]{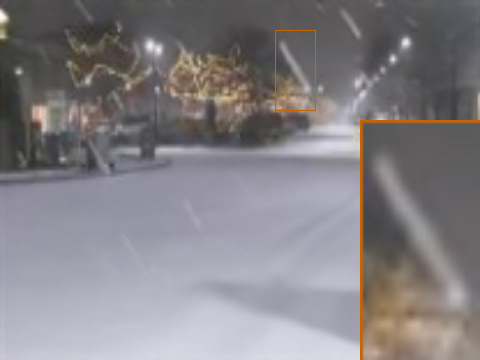}
	\includegraphics[height=0.74in]{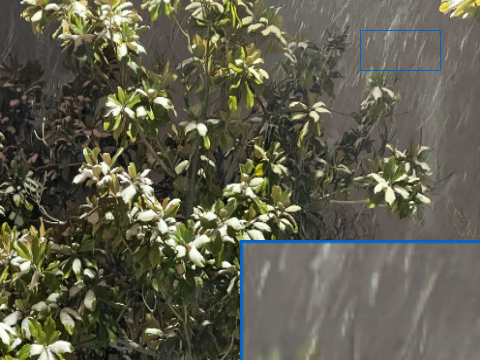}
	\includegraphics[height=0.74in]{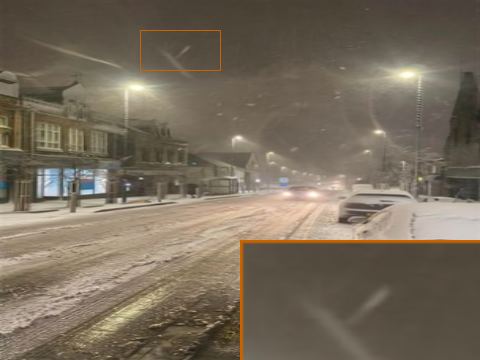}
	\includegraphics[height=0.74in]{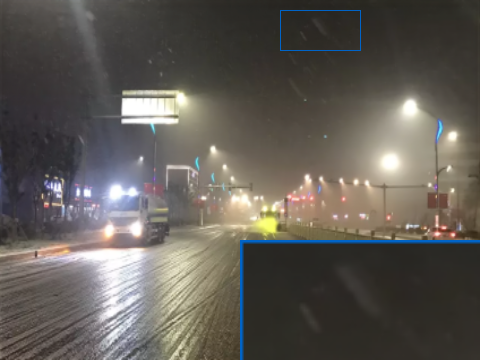}
	\includegraphics[height=0.74in]{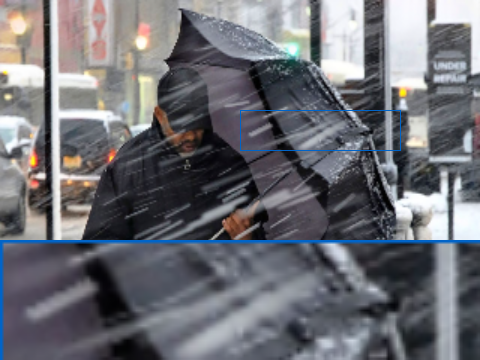}
	\includegraphics[height=0.74in]{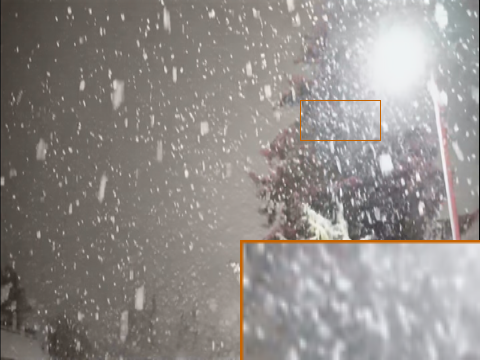}
	
	\vspace{1mm}
	\includegraphics[height=0.74in]{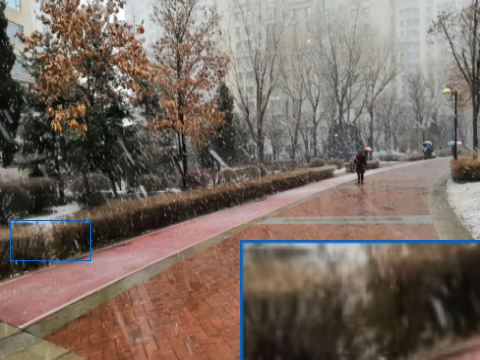}
	\includegraphics[height=0.74in]{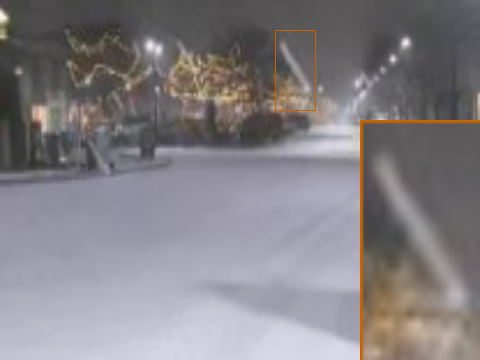}
	\includegraphics[height=0.74in]{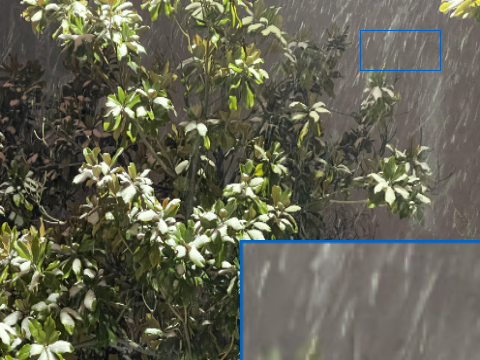}
	\includegraphics[height=0.74in]{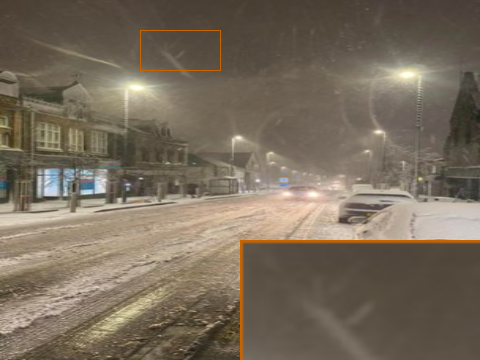}
	\includegraphics[height=0.74in]{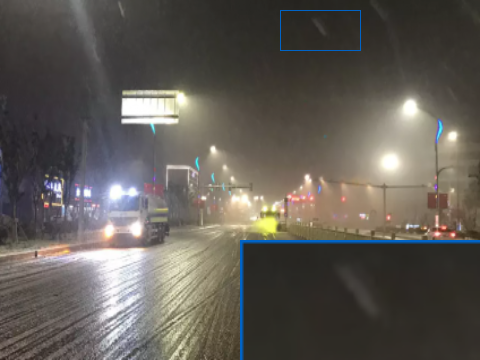}
	\includegraphics[height=0.74in]{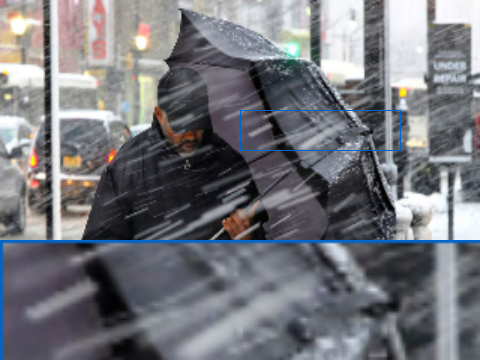}
	\includegraphics[height=0.74in]{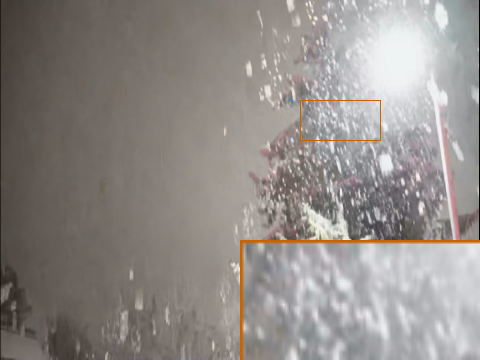}
	
	\vspace{1mm}
	\includegraphics[height=0.74in]{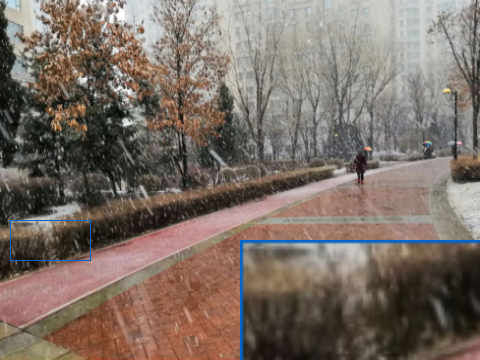}
	\includegraphics[height=0.74in]{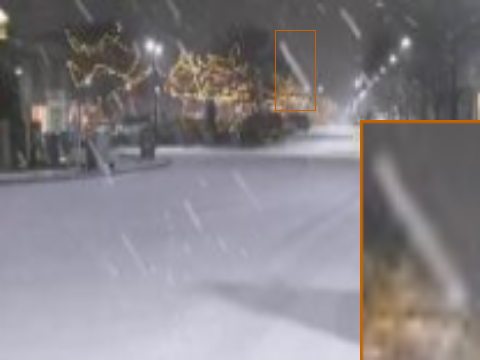}
	\includegraphics[height=0.74in]{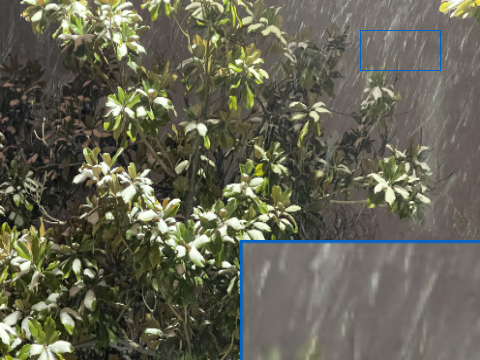}
	\includegraphics[height=0.74in]{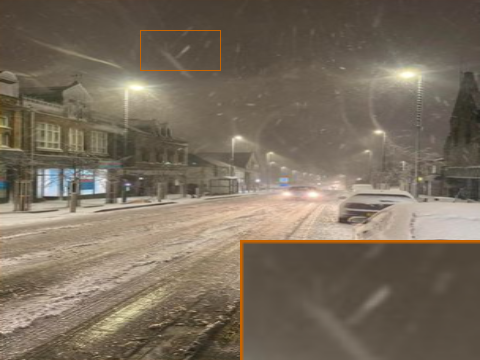}
	\includegraphics[height=0.74in]{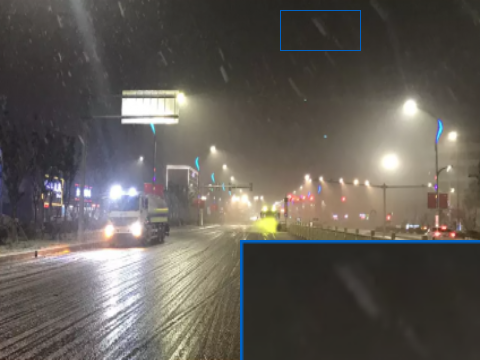}
	\includegraphics[height=0.74in]{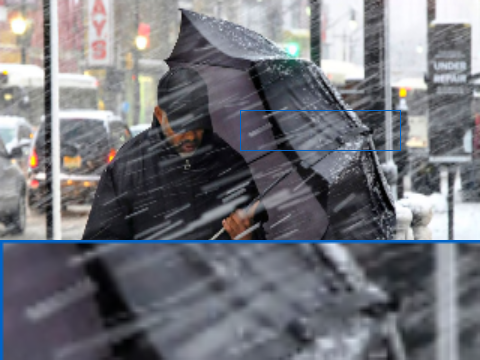}
	\includegraphics[height=0.74in]{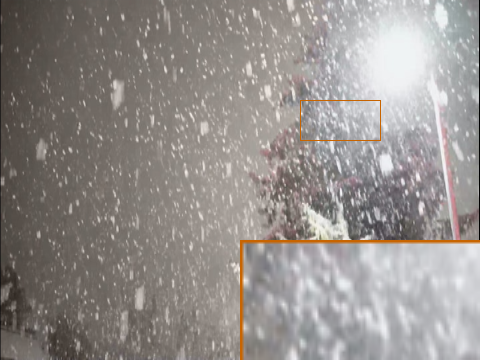}
	
	\vspace{1mm}
	\includegraphics[height=0.74in]{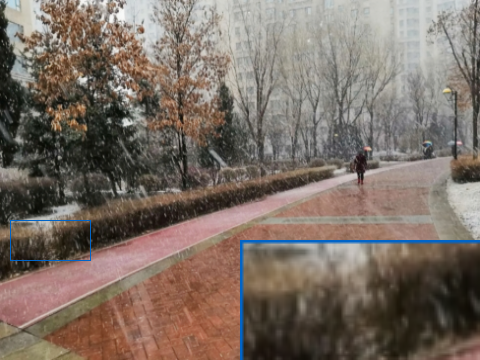}
	\includegraphics[height=0.74in]{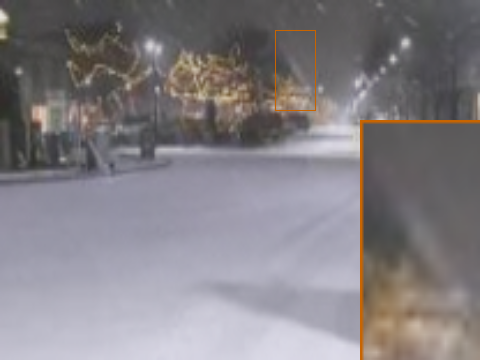}
	\includegraphics[height=0.74in]{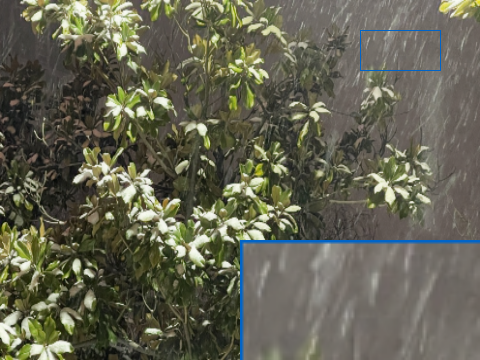}
	\includegraphics[height=0.74in]{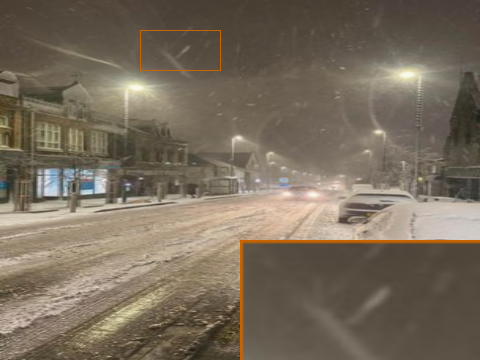}
	\includegraphics[height=0.74in]{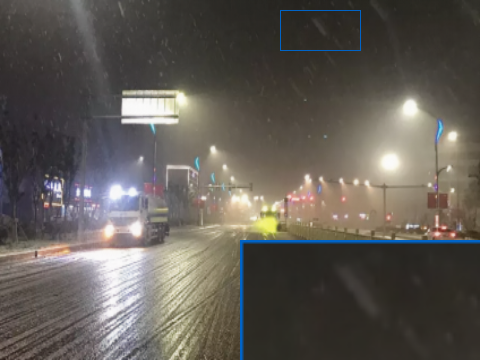}
	\includegraphics[height=0.74in]{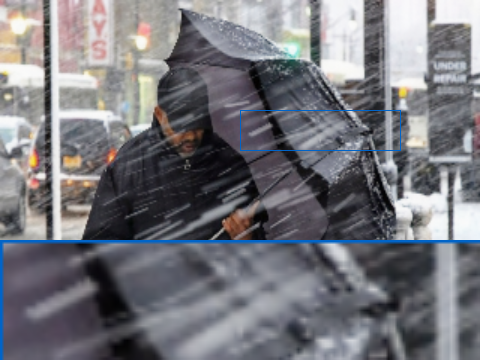}
	\includegraphics[height=0.74in]{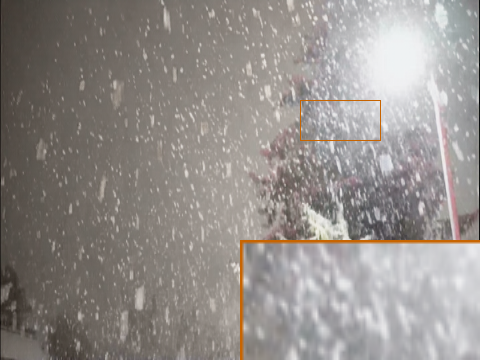}
	
	\vspace{1mm}
	\includegraphics[height=0.74in]{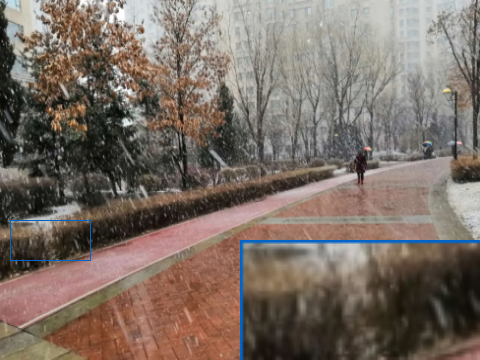}
	\includegraphics[height=0.74in]{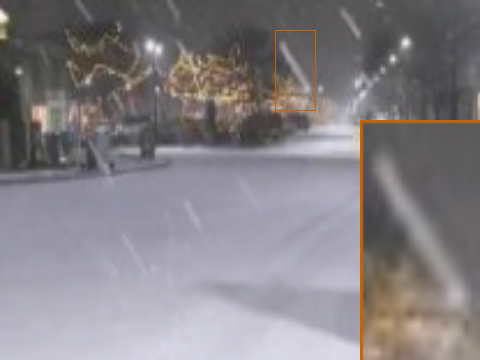}
	\includegraphics[height=0.74in]{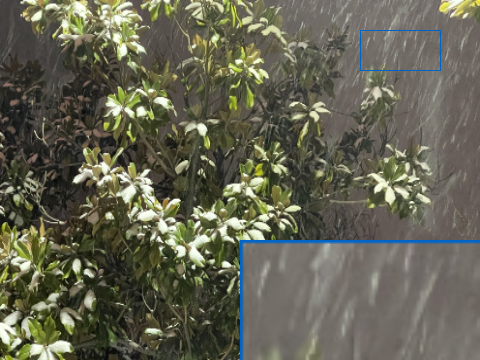}
	\includegraphics[height=0.74in]{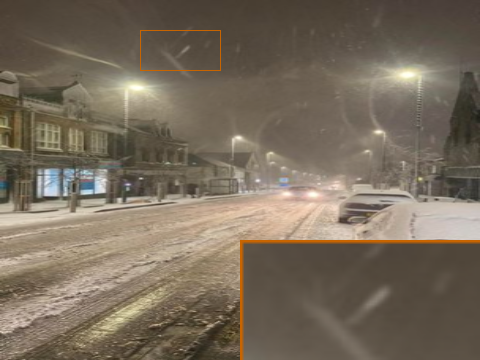}
	\includegraphics[height=0.74in]{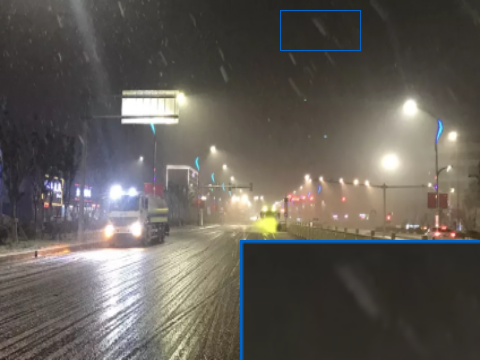}
	\includegraphics[height=0.74in]{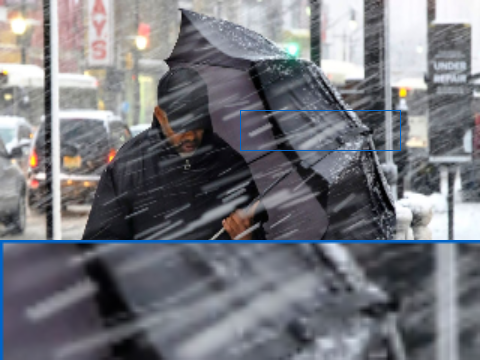}
	\includegraphics[height=0.74in]{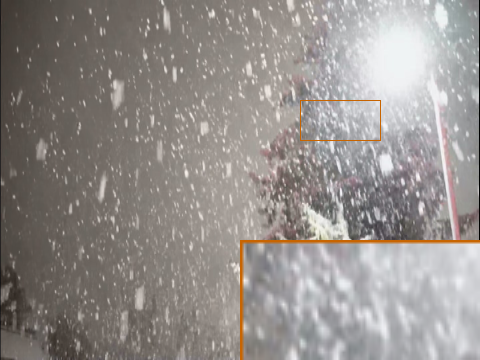}
	
	\vspace{1mm}
	\includegraphics[height=0.74in]{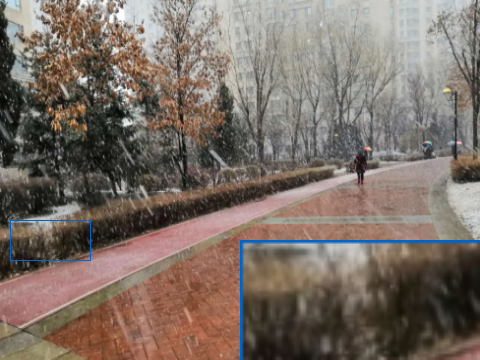}
	\includegraphics[height=0.74in]{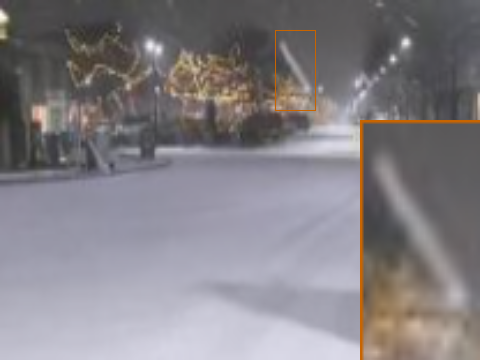}
	\includegraphics[height=0.74in]{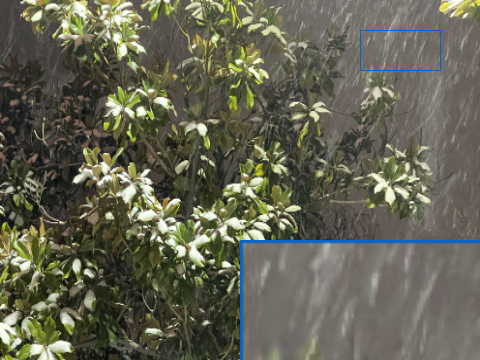}
	\includegraphics[height=0.74in]{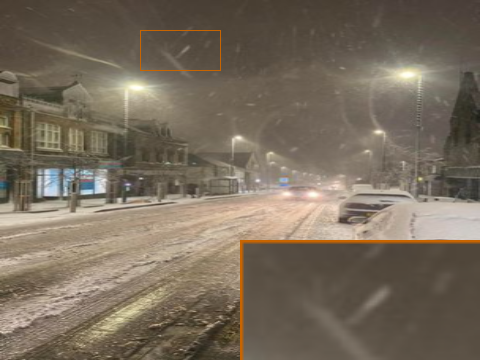}
	\includegraphics[height=0.74in]{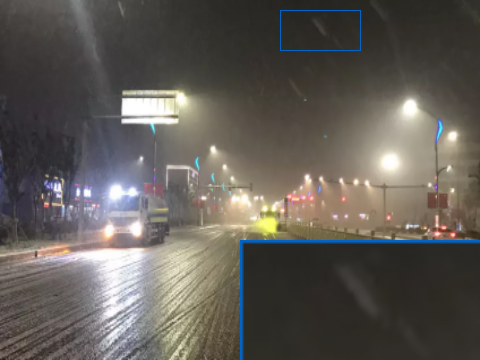}
	\includegraphics[height=0.74in]{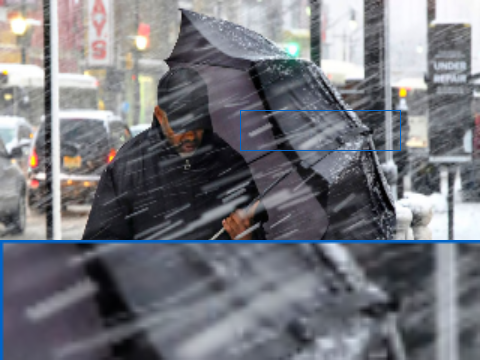}
	\includegraphics[height=0.74in]{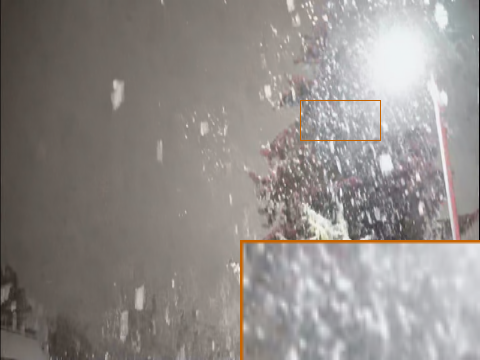}
	
	\vspace{1mm}
	\includegraphics[height=0.74in]{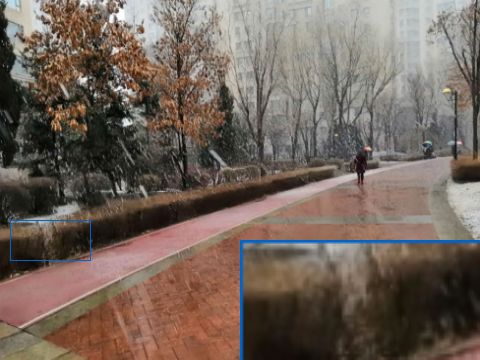}
	\includegraphics[height=0.74in]{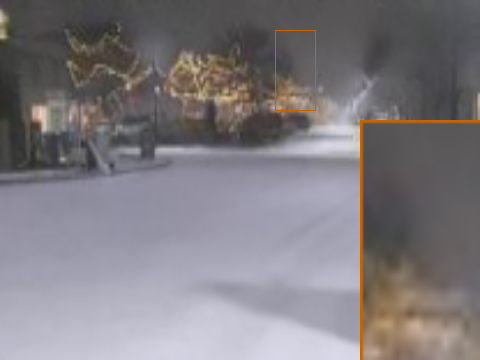}
	\includegraphics[height=0.74in]{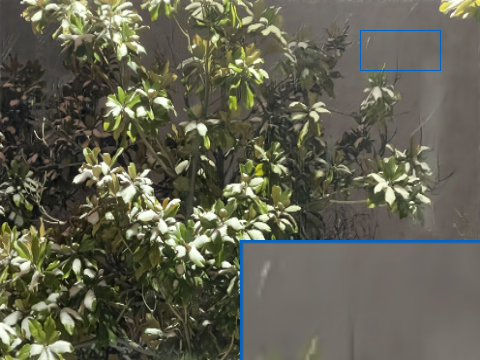}
	\includegraphics[height=0.74in]{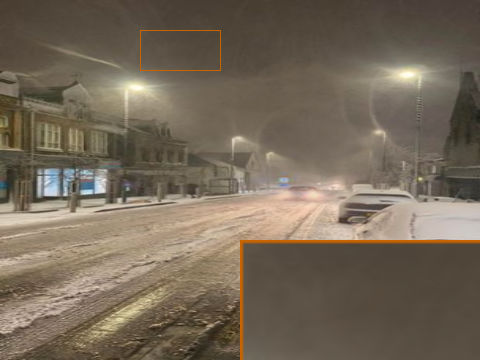}
	\includegraphics[height=0.74in]{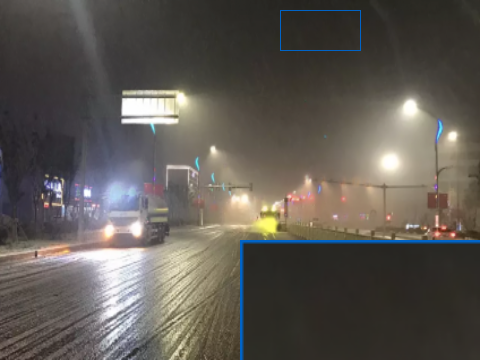}
	\includegraphics[height=0.74in]{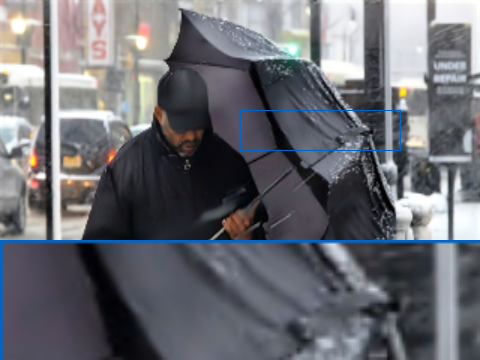}
	\includegraphics[height=0.74in]{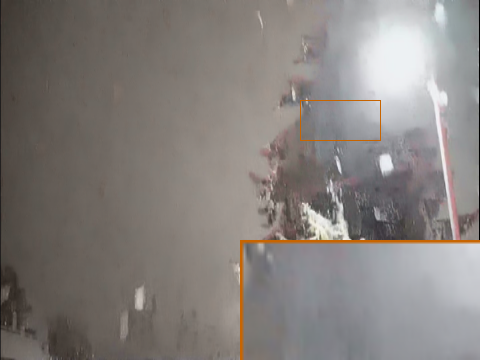}
	
	\caption{\textbf{Visual comparisons on RS300 dataset for real-world rain-by-snow weather removal}. From top to bottom: Input, DesnowNet \cite{liu2018desnownet}, PReNet\cite{ren2019progressive}, MPRNet\cite{zamir2021multi}, Uformer\cite{wang2022uformer}, TransWeather\cite{valanarasu2022transweather}, SnowFormer\cite{chen2022snowformer}, Restormer\cite{zamir2022restormer} and \textbf{RSFormer (ours)}.}
	\label{fig:rs300}
\end{figure*}

 \subsection{Loss Functions}
 Although the research on loss function is not the key in this paper, we still conduct ablation experiments to verify its effectiveness.
 As shown in TABLE \ref{tab:loss}, we compare $\mathcal{L}_1$, $\mathcal{L}_{spat}$ (Charbonnier \cite{charbonnier1994two}), $\mathcal{L}_{freq}$ and $\mathcal{L}_{spat}+\beta\cdot\mathcal{L}_{freq}$ losses.
 Overall, our proposed SFL loss obtains performance gains of 0.9 dB and 0.39 dB over commonly used $\mathcal{L}_1$ and $\mathcal{L}_{spat}$ loss.
 Meanwhile, compared to adopting frequency loss alone, the composition loss achieves a  better PSNR of 2.52 dB.
 Therefore, SFL loss comprehensively considers the similarity between spatial and frequency domains, thus improving the quality of the restored image.

 \begin{table}[H]
     \centering
     \caption{Ablation study of loss functions. The spatial-frequency loss utilized in our RSFormer achieves the highest score in PSNR.}
     \label{tab:loss}
     \renewcommand\arraystretch{1.25}
     \setlength{\tabcolsep}{8mm}{
     \begin{tabular}{lcc}
     \toprule
    Loss Function & PSNR & SSIM \\
    \midrule
    $\mathcal{L}_1$ & 33.64 & 0.952 \\
    $\mathcal{L}_{spat}$ & 34.15 & 0.969 \\
    $\mathcal{L}_{freq}$ & 32.02 & \textbf{0.984} \\
    $\mathcal{L}_{spat}+\beta\cdot\mathcal{L}_{freq}$ & \textbf{34.54} & 0.981  \\
    \bottomrule
    \end{tabular}
     }
 \end{table}

\subsection{Efficiency Analysis}

For real applications, floating point operation (FLOPs) (Billion), amount of parameters (Million) and inference time (second) are three significant factors. TABLE \ref{tab:efficiency} illustrates quantitative comparisons of recent efficient Restormer \cite{zamir2022restormer} and our proposed RSFormer, where inference time indicates the average required time to process 100 images with the size of 256$\times$256.
Our RSFormer increases FLOPs by 15.6\% and decreases the number of parameters by 1.53\%.
We also visualize the comparisons of inference time and corresponding PSNR metrics in Fig. \ref{fig:efficiency}.
Compared to Restormer \cite{zamir2022restormer}, our RSFormer achieves a better trade-off between the time consumption and quality of the output image. Therefore, our RSFormer performs both effectively and efficiently for rain-by-snow weather removal.

 \begin{table}[H]
     \centering
     \caption{Efficiency analysis of different methods and our proposed RSFormer. Our RSFormer requires the minimum time in inference process over Restormer.}
     \label{tab:efficiency}
     \renewcommand\arraystretch{1.25}
     \setlength{\tabcolsep}{4mm}{
     \begin{tabular}{lccc}
     \toprule
     Methods & FLOPs & Parameters & Inference Time \\
    \midrule
    Restormer\cite{zamir2022restormer}  & \textbf{141.0} & \underline{26.10} & \underline{0.282} \\
    \textbf{RSFormer}  & \underline{163.0} & \textbf{25.70} & \textbf{0.235} \\
    \bottomrule
    \end{tabular}
     } 
 \end{table}

 \subsection{Limitations}
 The main limitation of our approach lies in its reasonable inference time, although RSFormer has achieved a better trade-off than Restormer \cite{zamir2022restormer}.
 For a certain image with high resolution, our method can not even adopt GPU-accelerated calculation, which is one common problem of most Transformer-based image restoration methods \cite{wang2022uformer, zamir2022restormer, liang2021swinir}.
 Following the common designs, one stage of hierarchical architecture contains intra-stage learning and cross-stage sampling, we further correspondingly propose two suitable structures, named TCB and GLASM.
 However, as learning progresses, the intra-stage features do not remain consistent from beginning to end, yet we did not analyze and consider this progression.
 In addition, rain streaks and snow particles are added successively in our synthesized RSCityScape and RS100 datasets, whereas these two degradations appear simultaneously under real-world rain-by-snow weather condition with a certain correlation.

 \subsection{ConvNets or ViTs for Image Restoration?}
 Most current learning-based methods \cite{chen2021pre, zamir2022restormer, wang2022uformer, valanarasu2022transweather, chen2022snowformer, liang2021swinir} for image restoration are Transformer-based.
 However, recent work \cite{liu2022convnet, hou2022conv2former, ding2022scaling, liu2022more} either replace components of ViTs with convolution modules or modify ConvNets following the structure of ViTs in intra-stage feature learning, and demonstrate that ConvNets may achieve comparable or even superior performance over ViTs \cite{hou2022conv2former}.
 We explore the success of \cite{liu2021swin, yu2022metaformer, ding2022scaling, liang2021swinir, hou2022conv2former} and find that locality-wise global information act as a more significant part than globality-wise global information for intra-stage feature learning.
 Exactly, both window-based self-attention (inner product) and convolution perform well to model locality-wise global dependencies.
 Therefore, the only difference is that the properties of inner product and convolution lead to different response spaces, which is demonstrated in (\ref{eq:sa}) of self-attention, requiring three input to keep feature sizes, namely $Q$, $K$ and $V$, while convolution attention (c.f. (\ref{eq:ca})) only needs attention representation $A$ and value projection $V$.
 In addition to intra-stage feature learning, the other important component of hierarchical architecture is cross-stage progression.
 Our extensive experimental results demonstrate that both globality-wise and locality-wise global information are vital for the ultimate performance to remove rain-by-snow weather.
 Consequently, fine-tuned ConvNets can always outperform or approach ViTs without the consideration of cross-stage progression, and thereby our experiments suggest more attention to the cross-stage progression of hierarchical architectures for image restoration.

\section{Real-World Application}
To evaluate our RSFormer on real-world rain-by-snow weather removal, we utilize natural image quality evaluator (NIQE) \cite{mittal2012making}, neural image assessment (NIMA) \cite{talebi2018nima}, integrated local natural image quality evaluator (IL-NIQE) \cite{zhang2015feature} and spatial-spectral entropy-based quality (SSEQ) \cite{liu2014no} metrics, where a lower value corresponds to a higher quality of the restored image except for NIMA.
We find that all methods trained on RSCityScape dataset
suffer from severe brightness distortion, thus all restored images are generated by the pre-trained models on our synthesized RS100K dataset.
We report the quantitative comparisons in TABLE \ref{tab:real}.

Our proposed RSFormer achieves the best scores in NIMA, IL-NIQE and SSEQ metrics, while the second-best in NIQE.
Compared with the second-best Restormer \cite{zamir2022restormer} in synthetic quantitative results, RSFormer obtains significant performance gains of 7.87\% and 5.90\% in IL-NIQE/SSEQ.
For further visual comparison, we present the generated images from different methods on the RS300 dataset with real-world rain-by-snow in Fig. \ref{fig:rs300}.
Our RSFormer removes most rain streaks (seen in the 4th and 5th columns) and snow particles (seen in the 1st, 7th columns) while other methods show disability for real-world application.
RSFormer is also effective to remove snow streaks (the movement of snow particles) as shown in the 2nd and 6th columns of Fig. \ref{fig:rs300}.
Therefore, our RSFormer trained on rain-by-snow datasets can be easily transferred to snow streaks removal tasks without extra training and fine-tuning.

\begin{table}[H]
	\caption{Real-world rain-by-snow weather removal results. Our RSFormer achieves the best score in NIMA, IL-NIQE and SSEQ over other methods.\label{tab:real}}
	\centering
	\renewcommand\arraystretch{1.25}
	\setlength{\tabcolsep}{3.5mm}{
		\begin{tabular}{lcccc}
			\toprule
			Method & NIQE & NIMA & IL-NIQE & SSEQ \\
			\midrule
			DesnowNet \cite{liu2018desnownet} & 4.898 & 3.738 & 22.01 & 27.92 \\
			PReNet \cite{ren2019progressive} & 4.929 & 3.445 & \underline{21.93} & \underline{26.88} \\
			MPRNet \cite{zamir2021multi} & 5.104 & 3.892 & 22.91 & 28.68 \\
			Uformer \cite{wang2022uformer} & 5.108 & 3.926 & 22.99 & 28.66 \\
			TransWeather \cite{valanarasu2022transweather} & \textbf{4.882} & 4.073 & 22.04 & 26.95 \\
			SnowFormer \cite{chen2022snowformer} & 4.924 & \underline{4.167} & 22.18 & 27.44 \\
			Restormer \cite{zamir2022restormer} & 5.013 & 4.084 & 22.49 & 28.13 \\
			\textbf{RSFormer} & \underline{4.896} & \textbf{4.235} & \textbf{20.72} & \textbf{26.47} \\
			\bottomrule
		\end{tabular}
	}
\end{table}

\section{Conclusion}
In this paper, we focus on rain-by-snow weather removal, which is both practical and challenging image restoration task.
Overall, we propose an effective and efficient Transformer named RSFormer to eliminate coexisting rain streaks and snow particles.
Specifically, by exploring the proximity of intra-stage feature learning of ConvNets and ViTs in hierarchical architecture, we propose a Transformer-like convolution block to extract features instead of current calculation-heavy self-attention ViTs.
Meanwhile, we argue that cross-stage progression plays another significant role in performance improvement, which fails to be considered enough by most existing image restoration methods.
On this basis, we develop a global-local self-attention sampling mechanism (GLASM) to down-/up-sample features other than common-used (transposed) convolution or pixel-(un)shuffle.
Our GLASM captures both global and local dependencies for accurate sampling and information-rich feature extraction.
In addition, we synthesize two novel rain-by-snow datasets, RSCityScape and RS100K, to evaluate the performance of our RSFormer as benchmarks. We also collect a real-world rain-by-snow dataset for real application.
Extensive experiments verify the effectiveness and efficiency of our proposed RSFormer.

\vspace{-5mm}
\begin{IEEEbiography}[{\includegraphics[width=1in,height=1.25in,clip,keepaspectratio]{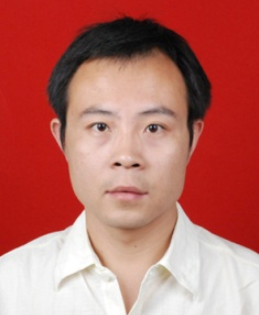}}]{Tao Gao}
(Member, IEEE)
received the B.Sc. degree, M.Sc. and Ph.D. degree in School of Electronics and Information, Northwestern Polytechnical University  in 2002, 2016 and 2010. He is a professor in School of Information Engineering, Chang'an University, Xi'an, China. His current research interests include image processing and computer vision.
\end{IEEEbiography}

\vspace{-5mm}
\begin{IEEEbiography}[{\includegraphics[width=1in,height=1.25in,clip,keepaspectratio]{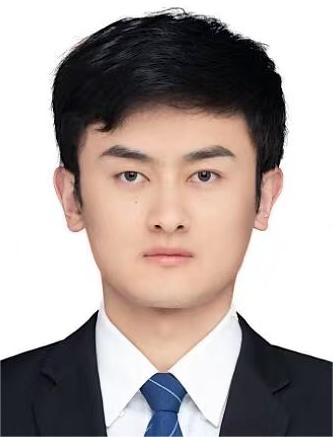}}]{Yuanbo Wen}
received B.Sc. degree in 2021 and is a Ph.D. candidate at present in School of Information Engineering, Chang'an University, Xi'an China. He is engaging in the research of computer vision and intelligent transportation. 
\end{IEEEbiography}

\vspace{-5mm}
\begin{IEEEbiography}[{\includegraphics[width=1in,height=1.25in,clip,keepaspectratio]{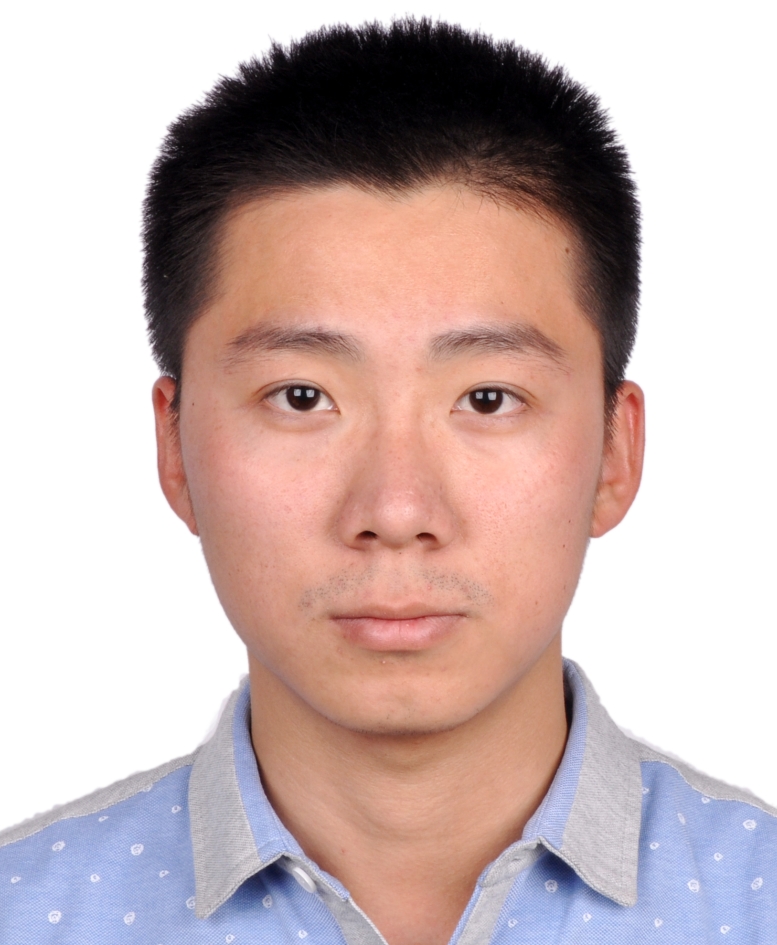}}]{Kaihao Zhang}
is currently pursuing the Ph.D. degree with the College of Engineering and Computer Science, The Australian National University, Canberra, ACT, Australia. His research interests focus on computer vision and deep learning. He has more than 20 referred publications in international conferences and journals, including CVPR, ICCV, ECCV, NeurIPS, AAAI, ACMMM, TPAMI, TIP, TMM, etc.
\end{IEEEbiography}

\vspace{-5mm}
\begin{IEEEbiography}[{\includegraphics[width=1in,height=1.25in,clip,keepaspectratio]{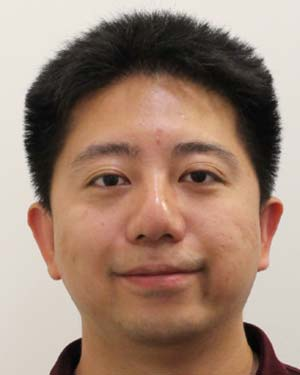}}]{Peng Cheng}
(Member, IEEE) received the Ph.D. degree in tele-communications from Shanghai JiaoTong University, Shanghai, China, in 2013.
He is currently a Senior Lecturer (Tenured Associate Professor in U.S. systems) with the Department of Computer Science and Information Technology, La Trobe University, Sydney, Australia, and is afﬁliated with the University of Sydney. His current research interests include wireless AI, machine learning, IoT, millimeter-wave communications, and compressive sensing theory.
\end{IEEEbiography}

\vspace{-5mm}
\begin{IEEEbiography}[{\includegraphics[width=1in,height=1.25in,clip,keepaspectratio]{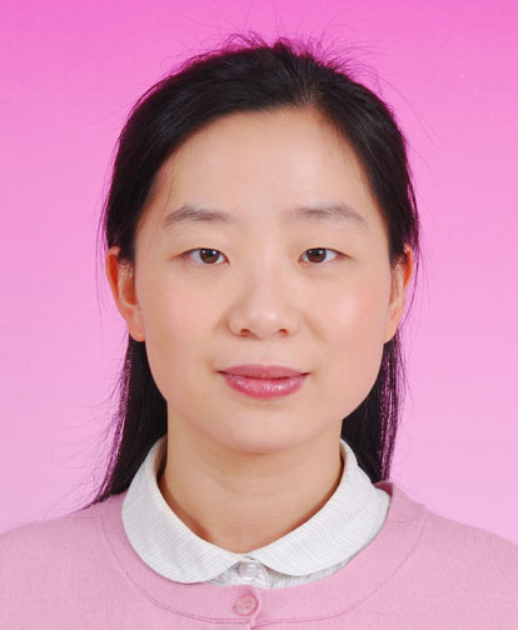}}]{Ting Chen}
 received the Ph.D. degree in Information and
Communication Engineering from Xidian University
in 2011. She is an assistant professor in School
of Information Engineering, Chang’an University,
Xi’an, China. Her research interests include image
processing, wireless network, \etc.
\end{IEEEbiography}

\vfill

\end{document}